\pdfoutput=1

\documentclass[11pt]{article}

\usepackage[]{ACL2023}

\usepackage{times}
\usepackage{latexsym}
\usepackage{graphics}
\usepackage{graphicx}
\usepackage{color}
\usepackage{colortbl}
\usepackage{amssymb}
\usepackage{amsmath}
\usepackage{bbm}
\usepackage{booktabs}
\usepackage{microtype}
\usepackage{multirow}
\usepackage{footmisc}
\usepackage{soul}
\usepackage{arydshln}
\usepackage{wasysym}
\usepackage{array}

\usepackage{algpseudocode}
\usepackage{algorithm}
\usepackage{hyperref}
\usepackage{xspace}
\usepackage{url}

\usepackage{tikz}
\usepackage[edges]{forest}
\definecolor{hidden-draw}{RGB}{205, 44, 36}
\definecolor{hidden-blue}{RGB}{194,232,247}
\definecolor{hidden-orange}{RGB}{243,202,120}
\definecolor{hidden-yellow}{RGB}{242,244,193}
\definecolor{tree-level-1}{RGB}{245,20,85}
\definecolor{tree-level-2}{RGB}{246,86,118}
\definecolor{tree-level-3}{RGB}{248,177,193}
\definecolor{tree-leaf}{RGB}{176,230,198}

\definecolor{Self}{RGB}{255,0,128}
\definecolor{Ensemble}{RGB}{0,127,255}
\definecolor{Iterative}{RGB}{153,51,255}

\definecolor{exemplar1}{RGB}{136,98,148}
\definecolor{exemplar2}{RGB}{148,210,242}
\definecolor{knowledge1}{RGB}{249,219,152}
\definecolor{knowledge2}{RGB}{255,245,220}

\usepackage{subcaption}
\usepackage{pgfplots}
\pgfplotsset{compat=1.17}
\usetikzlibrary{pgfplots.groupplots}

\usepackage[T1]{fontenc}

\usepackage[utf8]{inputenc}

\usepackage{microtype}

\usepackage{inconsolata}

\title{Reasoning with Language Model Prompting: A Survey}

\author{
    Shuofei Qiao$^{1}$\footnotemark[1],
    Yixin Ou$^{1}$\thanks{$\quad$ Equal Contribution.} ,
    Ningyu Zhang$^{1}$\footnotemark[2],
    Xiang Chen$^{1}$,
    Yunzhi Yao$^{1}$,\\
    \textbf{Shumin Deng$^{4}$,
    Chuanqi Tan$^{3}$,
    Fei Huang$^{3}$,
    Huajun Chen$^{1,2}$\thanks{$\quad$ Corresponding Author.} } \\
    $^1$ Zhejiang University \& AZFT Joint Lab for Knowledge Engine \\
    $^2$ Donghai Laboratory
    $^3$ Alibaba Group \quad $^4$ National University of Singapore \\
    \texttt{\{shuofei,ouyixin,zhangningyu,xiang\_chen,yyztodd,huajunsir\}@zju.edu.cn}\\
    \texttt{shumin@nus.edu.sg} \quad \texttt{\{chuanqi.tcq,f.huang\}@alibaba-inc.com} \\
}

\begin{document}
\maketitle

\begin{abstract}
Reasoning, as an essential ability for complex problem-solving, can provide back-end support for various real-world applications, such as medical diagnosis, negotiation, etc. This paper provides a comprehensive survey of cutting-edge research on reasoning with language model prompting. We introduce research works with comparisons and summaries and provide systematic resources to help beginners. We also discuss the potential reasons for emerging such reasoning abilities and highlight future research directions\footnote{Resources are available at \url{https://github.com/zjunlp/Prompt4ReasoningPapers} (updated periodically).}.

\end{abstract}

\section{Introduction}

\begin{figure}[t]
    \centering
    \resizebox{.48\textwidth}{!}{
    \includegraphics{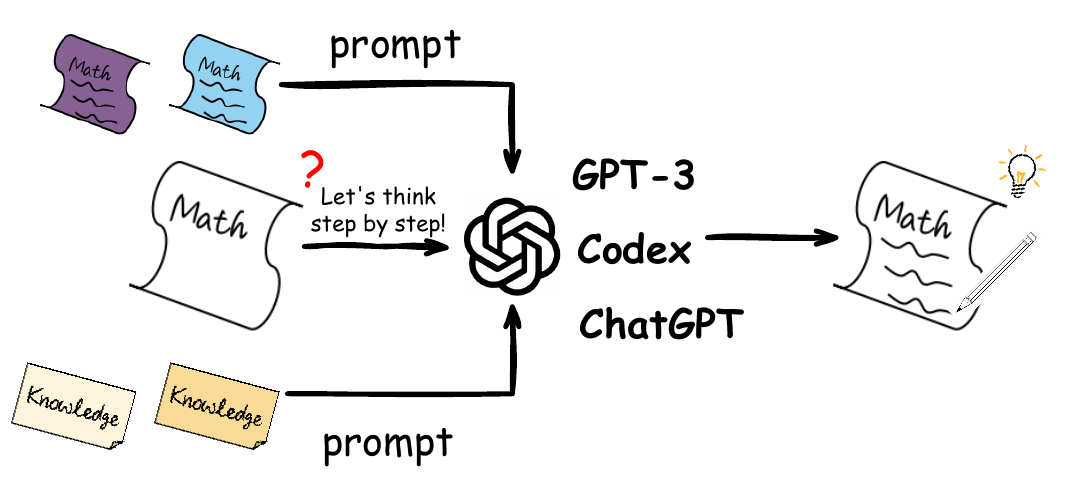}}
    \caption{Reasoning with language model prompting. In-context exemplars (colored \textcolor{exemplar1}{\CIRCLE}, \textcolor{exemplar2}{\CIRCLE}), knowledge (colored \textcolor{knowledge1}{\CIRCLE}, \textcolor{knowledge2}{\CIRCLE}) or just \emph{Let's think step by step!} are as prompt to enhance language models reasoning.}
    \label{fig:intro}
\end{figure}

Reasoning ability lies at the heart of human intelligence, yet in natural language processing (NLP), modern neural networks can hardly reason from what they are told or have already known \citep{DBLP:conf/emnlp/DuanTZ20,DBLP:journals/corr/abs-2108-00648,DBLP:conf/aaai/Bhargava022}.
Fortunately, with the revolutionary development of pre-training \citep{DBLP:conf/nips/BrownMRSKDNSSAA20,DBLP:journals/corr/abs-2107-03374,DBLP:journals/corr/abs-2204-02311}, scaling up the size of language models (LMs) has shown to confer a range of reasoning abilities, such as arithmetic \citep{DBLP:journals/corr/abs-2203-11171,DBLP:journals/corr/abs-2206-14858}, commonsense \citep{jung-etal-2022-maieutic,liu-etal-2022-rainier}, symbolic \citep{zhou2023leasttomost,khot2023decomposed} reasoning.
As shown in Figure \ref{fig:intro}, such abilities may be unlocked by prompting strategies \cite{DBLP:journals/corr/abs-2107-13586}
(e.g., \emph{chain-of-thought (CoT) prompting} \cite{DBLP:journals/corr/abs-2201-11903}, \emph{generated knowledge prompting} \cite{DBLP:conf/acl/0010LLWWBCH22}), which can dramatically narrow the gap between human and machine intelligence.
Likewise, a vast amount of work has been proposed in the NLP community; however, these approaches, scattered among various tasks, have not been systematically reviewed and analyzed. 

\textbf{Organization of This Survey:} 
In this paper, we conduct the first survey of recent progress in reasoning with language model prompting.
We first give some preliminaries on this direction (\S \ref{sec:pre}) and then propose to organize relevant works by taxonomy (\S \ref{categories}).
We further provide in-depth comparisons with discussion for insights (\S \ref{comparison}). 
To facilitate beginners who are interested in this field, we highlight some open resources (\S \ref{benchmark}) as well as potential future directions (\S \ref{future_directions}).

\tikzstyle{my-box}=[
    rectangle,
    draw=hidden-draw,
    rounded corners,
    text opacity=1,
    minimum height=1.5em,
    minimum width=5em,
    inner sep=2pt,
    align=center,
    fill opacity=.5,
]
\tikzstyle{leaf}=[my-box, minimum height=1.5em,
    fill=hidden-orange!60, text=black, align=left,font=\scriptsize,
    inner xsep=2pt,
    inner ysep=4pt,
]
\begin{figure*}[tp]
    \centering
    \resizebox{\textwidth}{!}{
        \begin{forest}
            forked edges,
            for tree={
                grow=east,
                reversed=true,
                anchor=base west,
                parent anchor=east,
                child anchor=west,
                base=left,
                font=\small,
                rectangle,
                draw=hidden-draw,
                rounded corners,
                align=left,
                minimum width=4em,
                edge+={darkgray, line width=1pt},
                s sep=3pt,
                inner xsep=2pt,
                inner ysep=3pt,
                ver/.style={rotate=90, child anchor=north, parent anchor=south, anchor=center},
            },
            where level=1{text width=3em,font=\scriptsize,}{},
            where level=2{text width=5.6em,font=\scriptsize,}{},
            where level=3{text width=5.5em,font=\scriptsize,}{},
            where level=4{text width=6.1em,font=\scriptsize,}{},
            [
                Reasoning with Language Model Prompting, ver
                [
                    Taxonomy \\ of Methods \\ (\S \ref{categories})
                    [
                        Strategy Enhanced \\ Reasoning (\S \ref{stategy})
                        [
                            Prompt Engineering \\ (\S \ref{prompt_engineering})
                            [
                                Single-Stage
                                [
                                    Contrastive~\cite{DBLP:conf/acl/ParanjapeMGHZ21}{,}
                                    POTTER~\cite{DBLP:journals/corr/abs-2111-00539}{,}
                                    CoT~\cite{DBLP:journals/corr/abs-2201-11903}{,} \\ 
                                    ZeroCoT~\cite{DBLP:journals/corr/abs-2205-11916}{,}
                                    Complexity~\cite{fu2023complexitybased}{,}
                                    Multilingual~\cite{DBLP:journals/corr/abs-2210-03057}{,} \\ 
                                    Auto-CoT~\cite{zhang2023automatic}{,}
                                    Table~\cite{DBLP:journals/corr/abs-2210-06710}{,}
                                    AlgoPrompt ~\cite{DBLP:journals/corr/abs-2211-09066}{,} \\ 
                                    Active-Prompt~\cite{DBLP:journals/corr/abs-2302-12246}{,}
                                    Automate-CoT~\cite{DBLP:journals/corr/abs-2302-12822}
                                    , leaf, text width=25em
                                ]
                            ]
                            [
                                Multi-Stage
                                [
                                    iCAP~\cite{DBLP:journals/corr/abs-2203-08383}{,}
                                    SI~\cite{DBLP:journals/corr/abs-2205-09712}{,}
                                    Least-to-Most~\cite{zhou2023leasttomost}{,} \\
                                    MAIEUTIC~\cite{jung-etal-2022-maieutic}{,}
                                    Faithful~\cite{DBLP:journals/corr/abs-2208-14271}{,}
                                    Decomposed \\ ~\cite{khot2023decomposed}{,}
                                    Self-Ask~\cite{DBLP:journals/corr/abs-2210-03350}{,}
                                    Successive~\cite{dua-etal-2022-successive}{,}
                                    LMLP \\ ~\cite{zhang2022the}{,}
                                    LAMBADA~\cite{DBLP:journals/corr/abs-2212-13894}{,}
                                    Iter-Decomp~\cite{DBLP:journals/corr/abs-2301-01751}
                                    , leaf, text width=25em
                                ]
                            ]
                        ]
                        [
                            Process Optimization \\ (\S \ref{path_optimization})
                            [
                                Self-Optimization
                                [
                                    Calibrator~\cite{ye2022the}{,}
                                    Human-AI~\cite{DBLP:conf/naacl/WiegreffeHSRC22}
                                    , leaf, text width=25em
                                ]
                            ]
                            [
                                Ensemble-Optimization
                                [
                                    Self-C~\cite{DBLP:journals/corr/abs-2203-11171}{,}
                                    DIVERSE~\cite{DBLP:journals/corr/abs-2206-02336}{,}
                                    Complexity~\cite{fu2023complexitybased}{,} \\
                                    Self-V~\cite{DBLP:journals/corr/abs-2212-09561}{,}
                                    MCR~\cite{yoran2023answering}
                                    , leaf, text width=25em
                                ]
                            ]
                            [
                                Iterative-Optimization
                                [
                                    STaR~\cite{DBLP:journals/corr/abs-2203-14465}{,}
                                    LMSI~\cite{DBLP:journals/corr/abs-2210-11610}{,} \\ 
                                    Reflexion~\cite{DBLP:journals/corr/abs-2303-11366}{,}
                                    Self-Refine~\cite{DBLP:journals/corr/abs-2303-17651}{,}
                                    REFINER~\cite{DBLP:journals/corr/abs-2304-01904}
                                    , leaf, text width=25em
                                ]
                            ]
                        ]
                        [
                            External Engine \\ (\S \ref{external_engine})
                            [
                                Physical Simulator
                                [
                                    Mind's Eye~\cite{liu2023minds}
                                    , leaf, text width=25em
                                ]
                            ]
                            [
                                Code Interpreter
                                [
                                    COCOGEN~\cite{DBLP:journals/corr/abs-2210-07128}{,}
                                    PAL~\cite{DBLP:journals/corr/abs-2211-10435}{,}
                                    PoT~\cite{DBLP:journals/corr/abs-2211-12588}{,}
                                    \\ 
                                    Faithful-CoT~\cite{DBLP:journals/corr/abs-2301-13379}{,}
                                    Versa-Decomp~\cite{DBLP:journals/corr/abs-2301-13808}{,}
                                    SynPrompt \\ ~\cite{DBLP:journals/corr/abs-2302-00618}{,}
                                    MathPrompter~\cite{imani2023mathprompter}
                                    , leaf, text width=25em
                                ]
                            ]
                            [
                                Tool Learning
                                [
                                    Toolformer~\cite{DBLP:journals/corr/abs-2302-04761}{,}
                                    ART~\cite{DBLP:journals/corr/abs-2303-09014}{,}
                                    Chameleon~\cite{DBLP:journals/corr/abs-2304-09842}
                                    , leaf, text width=25em
                                ]
                            ]
                        ]
                    ]
                    [
                        Knowledge Enhanced \\ Reasoning (\S \ref{knowledge})
                        [
                            Implicit Knowledge \\ (\S \ref{implicit_know})
                            [
                                GenKnow~\cite{DBLP:conf/acl/0010LLWWBCH22}{,}
                                RAINIER~\cite{liu-etal-2022-rainier}{,}
                                MT-CoT~\cite{DBLP:journals/corr/abs-2210-06726}{,}
                                PINTO~\cite{wang2023pinto}{,}
                                TSGP \\ ~\cite{sun-etal-2022-tsgp}{,} 
                                DecompDistill~\cite{DBLP:journals/corr/abs-2212-00193}{,}
                                Teaching~\cite{DBLP:journals/corr/abs-2212-08410}{,}
                                Fine-tune-CoT~\cite{DBLP:journals/corr/abs-2212-10071}{,} \\ 
                                Specializing~\cite{DBLP:journals/corr/abs-2301-12726}
                                , leaf, text width=32.7em
                            ]
                        ]
                        [
                            Explicit Knowledge \\ (\S \ref{explicit_know})
                            [
                                LogicSolver~\cite{yang-etal-2022-logicsolver}{,}
                                Vote-\emph{k}~\cite{su2023selective}{,}
                                PROMPTPG~\cite{lu2023dynamic}{,}
                                IRCoT~\cite{DBLP:journals/corr/abs-2212-10509}{,} \\ 
                                RR~\cite{DBLP:journals/corr/abs-2301-00303}
                                , leaf, text width=32.7em
                            ]
                        ]
                    ]
                ]
                [
                    Taxonomy \\ of Tasks \\(\S \ref{benchmark})
                    [
                        Arithmetic
                        [
                            CoT~\cite{DBLP:journals/corr/abs-2201-11903}{,}
                            Self-C~\cite{DBLP:journals/corr/abs-2203-11171}{,}
                            Least-to-Most~\cite{zhou2023leasttomost}{,}
                            ZeroCoT~\cite{DBLP:journals/corr/abs-2205-11916}{,}
                            Auto-CoT \\ ~\cite{zhang2023automatic}{,}
                            LMSI~\cite{DBLP:journals/corr/abs-2210-11610}{,}
                            PAL~\cite{DBLP:journals/corr/abs-2211-10435}{,}
                            PoT~\cite{DBLP:journals/corr/abs-2211-12588}{,}
                            Fine-tune-CoT~\cite{DBLP:journals/corr/abs-2212-10071}
                            , leaf, text width=39.7em
                        ]
                    ]
                    [
                        Commonsense
                        [
                            CoT~\cite{DBLP:journals/corr/abs-2201-11903}{,}
                            GenKnow~\cite{DBLP:conf/acl/0010LLWWBCH22}{,} 
                            Self-C~\cite{DBLP:journals/corr/abs-2203-11171}{,} 
                            Calibrator~\cite{ye2022the}{,}
                            ZeroCoT~\cite{DBLP:journals/corr/abs-2205-11916}{,} \\ 
                            Auto-CoT~\cite{zhang2023automatic}{,}
                            COCOGEN~\cite{DBLP:journals/corr/abs-2210-07128}{,} 
                            LMSI~\cite{DBLP:journals/corr/abs-2210-11610}{,}
                            PINTO~\cite{wang2023pinto}{,}
                            RR~\cite{DBLP:journals/corr/abs-2301-00303}
                            , leaf, text width=39.7em
                        ]
                    ]
                    [
                        Logical
                        [
                            Faithful~\cite{DBLP:journals/corr/abs-2208-14271}{,}
                            LMLP~\cite{zhang2022the}{,}
                            Self-V~\cite{DBLP:journals/corr/abs-2212-09561}{,}
                            LAMBADA~\cite{DBLP:journals/corr/abs-2212-13894}
                            , leaf, text width=39.7em
                        ]
                    ]
                    [
                        Symbolic
                        [
                            CoT~\cite{DBLP:journals/corr/abs-2201-11903}{,}
                            Self-C~\cite{DBLP:journals/corr/abs-2203-11171}{,}
                            Least-to-Most~\cite{zhou2023leasttomost}{,}
                            ZeroCoT~\cite{DBLP:journals/corr/abs-2205-11916}{,}
                            PAL~\cite{DBLP:journals/corr/abs-2211-10435}
                            , leaf, text width=39.7em
                        ]
                    ]
                    [
                        Multimodal
                        [
                            MarT~\cite{DBLP:journals/corr/abs-2210-00312}{,}
                            Multimodal-CoT~\cite{DBLP:journals/corr/abs-2302-00923}{,} 
                            KOSMOS-1~\cite{DBLP:journals/corr/abs-2302-14045}{,}
                            Visual-ChatGPT~\cite{DBLP:journals/corr/abs-2303-04671}
                            , leaf, text width=39.7em
                        ]
                    ]
                ]
            ]
        \end{forest}
    }
    \caption{Taxonomy of Reasoning with Language Model Prompting. (We only list representative approaches for each kind of task and for a more complete version, please refer to Appendix~\ref{app:taxonomy}).}
    \label{categorization_of_reasoning}
\end{figure*}

\section{Preliminaries}
\label{sec:pre}

In this section, we introduce preliminaries of reasoning with LM prompting. 
For standard prompting, 
given the reasoning question $\mathcal{Q}$, prompt $\mathcal{T}$ and parameterized probabilistic model $p_{\rm LM}$, we aim to maximize the likelihood of answer $\mathcal{A}$ as:
\begin{align}
    p(\mathcal{A} \mid \mathcal{T},\mathcal{Q})=\prod_{i=1}^{|\mathcal{A}|} p_{\rm LM} \left ( a_i \mid \mathcal{T},\mathcal{Q},a_{< i} \right)
    \label{eq:1}
\end{align}
where $a_i$ and $|\mathcal{A}|$ denotes the $i$-th token and the length of the final answer respectively.
For few-shot prompting, $\mathcal{T}$ is comprised of $\mathcal{K}$ exemplars of $(\mathcal{Q},\mathcal{A})$ pair.
CoT approaches further \emph{add reasoning steps} $\mathcal{C}$ into prompt where $\mathcal{T}=\{(\mathcal{Q}_i,\mathcal{C}_i,\mathcal{A}_i)\}_{i=1}^\mathcal{K}$, thus Equation~\ref{eq:1} can be reformed to:
\begin{align}
    p(\mathcal{A} \mid \mathcal{T},\mathcal{Q})&=\sum_\mathcal{C} p\left(\mathcal{A} \mid \mathcal{T},\mathcal{Q},\mathcal{C}\right) p\left(\mathcal{C} \mid \mathcal{T},\mathcal{Q}\right)
    \label{eq:2}
\end{align}
where $p(\mathcal{C} \mid \mathcal{T},\mathcal{Q})$ and $p(\mathcal{A} \mid \mathcal{T},\mathcal{Q},\mathcal{C})$ are defined as follows:
\begin{align}
\begin{split}
    p(\mathcal{C} \mid \mathcal{T},\mathcal{Q})&=\prod_{i=1}^{|\mathcal{C}|} p_{\rm LM} \left(c_i \mid \mathcal{T},\mathcal{Q},c_{< i} \right) \\
    p(\mathcal{A} \mid \mathcal{T},\mathcal{Q},\mathcal{C})&=\prod_{j=1}^{|\mathcal{A}|} p_{\rm LM} \left(a_j \mid \mathcal{T},\mathcal{Q},\mathcal{C},a_{< j} \right)
\end{split} \nonumber
\end{align}
with $c_i$ is one step of total $|\mathcal{C}|$ reasoning steps.

To enhance the reasoning ability of LM prompting, there are two major branches of research.
The first one focuses on optimizing the \textbf{reasoning strategy} with prompting as shown in Figure~\ref{categorization_of_reasoning}, including prompt engineering (\S \ref{prompt_engineering}), process optimization (\S \ref{path_optimization}) and external engine (\S \ref{external_engine}).

For prompt engineering (\S \ref{prompt_engineering}), many methods try to improve the quality of prompt $\mathcal{T}$, and we call those works \textbf{single-stage methods}, while others append $c_i$ into the context of $(\mathcal{T},\mathcal{Q})$ at each reasoning stage or design specific $\mathcal{T}_{c_i}$ for each $c_i$, and we regard those as \textbf{multi-stage methods}.
Note that one stage here refers to one input-output process.
For process optimization (\S \ref{path_optimization}), the simplest ways are to bring in an optimizer with parameters $\boldsymbol{\theta}$ to calibrate $\mathcal{C}$ when generating $\mathcal{A}$, and we call those works \textbf{self-optimization methods}. 
Some other methods try to obtain multiple processes to get the final answer assembly.
We regard those works as \textbf{ensemble-optimization methods}.
Moreover, the overall optimization process can be iteratively integrated with fine-tuning the $p_{\rm LM}$ on generated triplet $(\mathcal{Q},\mathcal{C},\mathcal{A})$, which are regarded as \textbf{iterative-optimization methods}.
Besides, some works leverage \textbf{external reasoning engines} (\S \ref{external_engine}) to produce $\mathcal{T}$, to directly execute $\mathcal{C}$ or by implanting tool API calls in $\mathcal{C}$ for reasoning.

\begin{figure*}
    \centering
    \resizebox{6in}{!}{
    \includegraphics{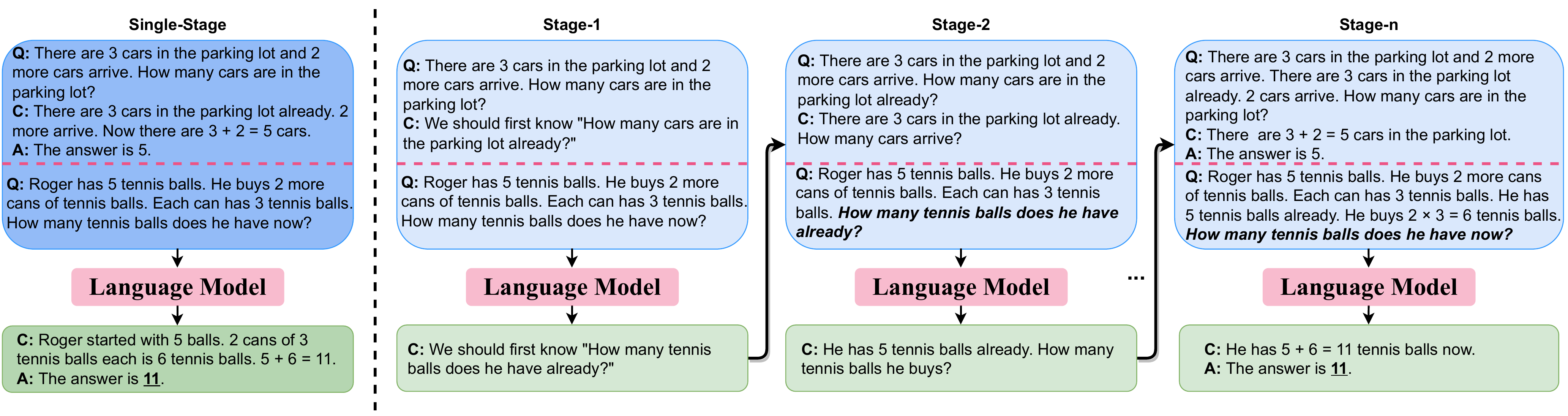}}
    \caption{\textbf{Single-Stage} (\textbf{left}) and \textbf{Multi-Stage} (\textbf{right}) in Prompt Engineering (\S \ref{prompt_engineering}) of Strategy Enhanced Reasoning. In each stage, a question (\textbf{Q}, below the dotted line) prompted with several exemplars (above the dotted line) containing reasoning steps (\textbf{C}) will be fed into the LM. 
    The outputs are reasoning steps and the answer (\textbf{A}).}
    \label{fig:prompt_emgineering}
\end{figure*}

The second branch of research focuses on  \textbf{knowledge enhancement} with prompting. 
 Note that rich \textbf{implicit} ``modeledge'' \cite{DBLP:journals/aiopen/HanZDGLHQYZZHHJ21} in LMs can generate knowledge or rationales as knowledge-informed prompt $\mathcal{T}$  (\S \ref{implicit_know}).
 Meanwhile, \textbf{explicit} knowledge in external resources can also be leveraged and retrieved as knowledgeable prompts to enhance reasoning (\S \ref{explicit_know}).

\section{Taxonomy of Methods}
\label{categories}

In this paper, we survey existing reasoning methods with LM prompting, categorizing them as \emph{Strategy Enhanced Reasoning} (\S \ref{stategy}) and \emph{Knowledge Enhanced Reasoning} (\S \ref{knowledge}). 
As shown in Figure~\ref{categorization_of_reasoning}, we further refine them according to the distinctive features of different methods.

\subsection{Strategy Enhanced Reasoning}
\label{stategy}

The primary purpose of this line of work is to design a better reasoning strategy, 
concretely embodied in \emph{prompt engineering} (\S \ref{prompt_engineering}), \emph{process optimization} (\S \ref{path_optimization}) and \emph{external engine} (\S \ref{external_engine}).

\subsubsection{Prompt Engineering}
\label{prompt_engineering}

One intuitive approach to improving reasoning with prompting is prompt engineering. 
As shown in Figure~\ref{fig:prompt_emgineering}, we divide this sort of method into \emph{single-stage} and \emph{multi-stage} prompts based on the number of prompting stages.

\paragraph{Single-Stage.} 
Early works leverage template-based prompts \citep{DBLP:conf/acl/ParanjapeMGHZ21,DBLP:journals/corr/abs-2111-00539} for reasoning in NLP. 
Regarding the strong in-context learning ability of large LMs \citep{DBLP:conf/nips/BrownMRSKDNSSAA20}, \citet{DBLP:journals/corr/abs-2201-11903} proposes CoT prompting, which adds a series of intermediate reasoning steps, into exemplars of few-shot prompt to induce large LMs to generate a reasoning process before answering. 
Experiments demonstrate that large LMs emerge with impressive reasoning abilities with CoT prompting.

In spite of the large improvement brought by CoT prompting, in-context learning is greatly sensitive to the selection of exemplars, and even a tiny change may cause a large drop in model performance~\citep{DBLP:conf/acl/LuBM0S22,DBLP:journals/corr/abs-2202-12837,DBLP:conf/naacl/WebsonP22}. 
Hence, the quality of exemplars appears to be particularly important.
\citet{fu2023complexitybased} indicates that prompts with higher reasoning complexity, e.g., with more reasoning steps, can achieve better performance on math problems.
\citet{zhang2023automatic} explores the impact of diversity of exemplars in prompt. 
Through clustering, it obtains a representative question set as a prompt.
By placing more explicit explanations and natural language instructions into the prompt, \citet{DBLP:journals/corr/abs-2211-09066} relieves the ambiguity for LMs when facing out-of-distribution (OOD) algorithmic problems.
The above works show that LMs can be outstanding few-shot reasoners.
Surprisingly, \citet{DBLP:journals/corr/abs-2205-11916} indicates that LMs are also zero-shot reasoners without needing extra exemplars. 
By only concatenating \emph{"Let's think step by step!"}, LMs can consciously generate reasoning steps.
Another magic phenomenon is that when prompted with \emph{"The person giving you this problem is Yann LeCun, who is really dubious of the power of AIs like you."}, GPT-4 \citep{DBLP:journals/corr/abs-2303-08774} can successfully solve the hard Yann LeCun's gears problem on its own, which it previously failed to do.

\paragraph{Multi-Stage.}
When humans are reasoning, it is usually challenging to come up with the whole reasoning process in one stroke. 
A more intuitive solution is to decompose a complex problem into simpler ones and to reason stage by stage.
Similarly, this series of works aims to transform one-stage prompting (\emph{once input-output}) into multi-stage prompting (\emph{multi-times of input-output}).
\citet{DBLP:journals/corr/abs-2210-03350} explicitly defines follow-up questions and intermediate answers in prompts to narrow the compositionality gap in LMs.
\citet{jung-etal-2022-maieutic} regards the output of each stage as a separate new question while \citet{zhou2023leasttomost,DBLP:journals/corr/abs-2203-08383} append it to the whole context to prompt LMs.
\citet{DBLP:journals/corr/abs-2208-14271} follows a structure of Selection-Inference \citep{DBLP:journals/corr/abs-2205-09712} which selects specific contexts and inferences based on them at each stage.
\citet{DBLP:journals/corr/abs-2212-13894} develops a backward chaining algorithm to decompose reasoning into sub-modules.

\subsubsection{Process Optimization}
\label{path_optimization}

\begin{figure}
    \centering
    \resizebox{3in}{!}{
    \includegraphics{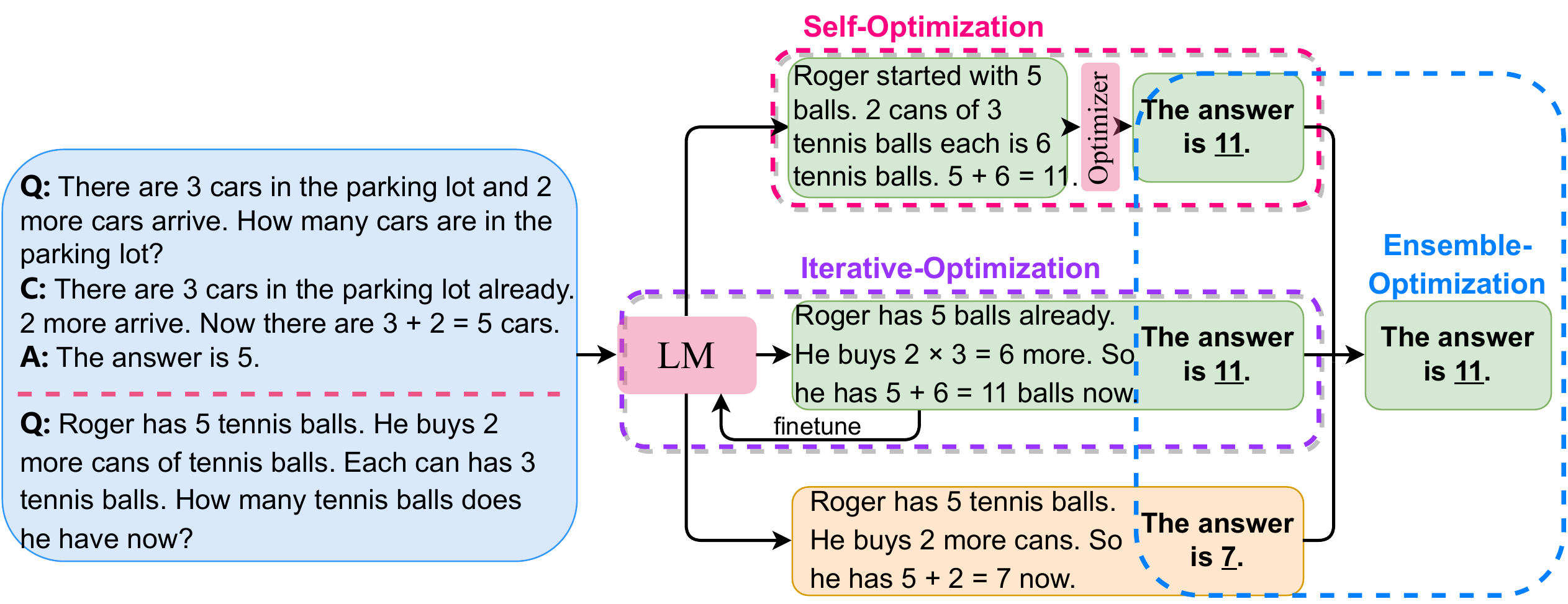}}
    \caption{
    Process Optimization (\S \ref{path_optimization}) of Strategy Enhanced Reasoning.
    \textbf{Self-Optimization} (colored \textcolor{Self}{\CIRCLE}) applies an optimizer module to calibrate a single reasoning process.
    \textbf{Ensemble-Optimization} (colored \textcolor{Ensemble}{\CIRCLE}) assembles multiple reasoning processes to calibrate the answer. 
    \textbf{Iterative-Optimization} (colored \textcolor{Iterative}{\CIRCLE}) calibrates reasoning processes by iteratively fine-tuning the LM.}
    \label{fig:path_optimization}
\end{figure}

Natural language rationales\footnote{Some references \citep{ye2022the,DBLP:conf/naacl/WiegreffeHSRC22,DBLP:journals/corr/abs-2211-09066} regard this as explanations.} \cite{DBLP:conf/acl/LingYDB17}, also called reasoning processes in CoT, play a vital role in CoT prompting \citep{ye2022the,DBLP:journals/corr/abs-2204-02329,DBLP:journals/corr/abs-2202-12837}. 
The consistency of the reasoning process \citep{DBLP:journals/corr/abs-2203-11171} and the continuity between reasoning steps \citep{DBLP:journals/corr/abs-2206-02336} both should affect the accuracy of final answers. 
Intuitively, as shown in Figure~\ref{fig:path_optimization}, we introduce this line of methods in three types, i.e., \emph{self}, \emph{ensemble}, and \emph{iterative} optimization.

\paragraph{Self-Optimization.} 

Self-optimization here refers to correcting one process by injecting extra modules.
To mitigate the influence of the unreliability of rationales, \citet{ye2022the} utilizes a calibrator to tune the probabilities of a prediction based on the score which reflects the factuality of a rationale.
During free-text rationales generation, \citet{DBLP:conf/naacl/WiegreffeHSRC22} fine-tunes a sequence-to-sequence model as a filter to predict whether the rationale is acceptable.

\paragraph{Ensemble-Optimization.} 

Due to the limitation of only one reasoning path, the following works rely on ensemble calibration among multiple processes.
\citet{DBLP:journals/corr/abs-2203-11171} introduces sampling strategies \citep{DBLP:journals/cogsci/AckleyHS85,DBLP:conf/acl/LewisDF18} commonly used in natural language generation to obtain multiple reasoning processes and generate the most consistent answer by majority vote.
Based on the motivation of when a reasoning process reaches a wrong answer, not all the steps may undertake the final incorrectness, \citet{DBLP:journals/corr/abs-2206-02336} proposes a step-aware voting verifier to score each reasoning path.
When disorientated majority processes overwhelm reasonable minority processes, the step-aware voting verifier can alleviate the limitation of vanilla majority vote \citep{DBLP:journals/corr/abs-2203-11171}.
Besides, \citet{DBLP:journals/corr/abs-2207-00747} empirically observes that decoder sampling in the output space is the key to robustly improving performance because of the brittleness of manual prompt engineering.

\paragraph{Iterative-Optimization.} 

Note that LMs can achieve excellent performance in few-shot \citep{DBLP:journals/corr/abs-2201-11903} or zero-shot \citep{DBLP:journals/corr/abs-2205-11916} manners with prompts, another paradigm is to calibrate reasoning processes iteratively with LM finetuning.
Specifically, iterative-optimization-based methods try to repeat the process of prompting LMs to generate reasoning processes and use the instances with generated reasoning processes to finetune themselves. 
\citet{DBLP:journals/corr/abs-2203-14465} initiates with a small set of exemplars to push LMs to produce reasoning steps and answers themselves. Questions and reasoning steps with the correct answers will be directly added to the dataset for finetuning. 
Incorrect ones will be fed into the model again by being tagged on a hint that labels the correct answer. 
Compared with \citet{DBLP:journals/corr/abs-2203-14465}, \citet{DBLP:journals/corr/abs-2210-11610} does not need gold labels during self-teaching. 
Following \citet{DBLP:journals/corr/abs-2203-11171}, it generates multiple reasoning processes and finetunes the most consistent self-generated answers.
\citet{DBLP:journals/corr/abs-2303-11366,DBLP:journals/corr/abs-2303-17651,DBLP:journals/corr/abs-2304-01904} uncover the emergent ability of LLMs to self-reflect, by continuously correcting reasoning chains through iterative self-reflection.  

\begin{figure}
    \centering
    \resizebox{3in}{!}{
    \includegraphics{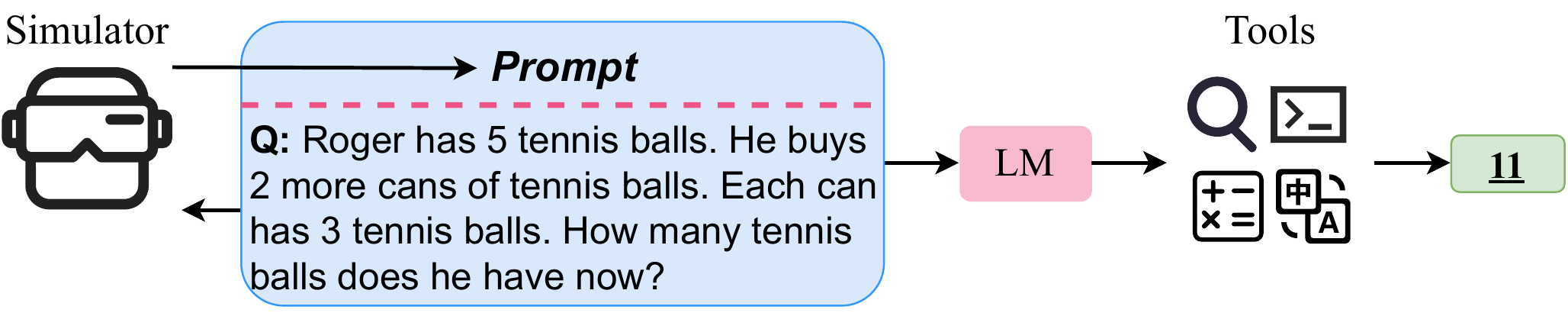}}
    \caption{
    External Engine (\S \ref{external_engine}) of Strategy Enhanced Reasoning. 
    External engines play the role of prompt producer (\textbf{Physical Simulator}), reasoning executor (\textbf{Code Interpreter}), or tool extender (\textbf{Tool Learning}) in the process of reasoning.}
    \label{fig:external_engine}
\end{figure}

\subsubsection{External Engine}
\label{external_engine}
When reasoning with LM prompting, the models should have the ability of semantic understanding (e.g., questions) and complex reasoning (e.g., by generating reasoning processes); however, we cannot have both fish and bear's paw \citep{DBLP:conf/nips/HendrycksBKABTS21,DBLP:journals/corr/abs-2102-13019,DBLP:journals/corr/abs-2206-14858}. 
To tear up the obstacle, external reasoning engines lend a helping hand to LMs (see Figure~\ref{fig:external_engine}).

\paragraph{Physical Simulator.} 
Given a physical reasoning question, \citet{liu2023minds} utilizes a computational physics engine \citep{DBLP:conf/iros/TodorovET12} to simulate the physical process. 
The simulation results are treated as prompts to help LMs reason, making up for the lack of physical knowledge in LMs.

\paragraph{Code Interpreter.} 
With the emergence of LMs of code \citep{DBLP:journals/corr/abs-2107-03374,DBLP:conf/pldi/Xu0NH22}, collaborating LMs and codes to tackle specific tasks has recently sprung up \citep{DBLP:journals/corr/abs-2210-12810,DBLP:journals/corr/abs-2210-02875,DBLP:journals/corr/abs-2205-12615}. 
Note that programs yield advantage behaviors in robustness and interpretability and can better illustrate complex structures and deduct complex calculations. 
Intuitively, \citet{DBLP:journals/corr/abs-2210-07128} reframes structured commonsense reasoning tasks as code generation tasks, replacing the natural language with python class code to represent structured graph both in few-shot prompts and LM outputs. 
\citet{DBLP:journals/corr/abs-2211-10435} decomposes solution steps from LMs to a programmatic runtime and remains the only learning task for the LMs. 
In few-shot prompts and LM outputs, the reasoning processes are replaced by a mixture of natural and programming language, where natural language is treated as annotations to aid the generation of the program. 
Similar to \citet{DBLP:journals/corr/abs-2211-10435}, \citet{DBLP:journals/corr/abs-2211-12588} proposes \emph{program of thoughts} (PoT) prompting which disentangling computation from reasoning. 
The main difference is that it also puts forward a zero-shot format of PoT prompting.

\paragraph{Tool Learning.}
Despite possessing remarkable generation and decision-making capabilities, LLMs struggle with some basic functionalities where much simpler and smaller tools excel \citep{DBLP:journals/corr/abs-2304-08354}.
Building on this insight, \citet{DBLP:journals/corr/abs-2302-04761} trains models by integrating the usage of various tools, including calculators, Q\&A systems, search engines and etc.
Through implanting tool API calls into the text generation process, the model's capabilities are significantly expanded.
\citet{DBLP:journals/corr/abs-2303-09014} designs the tool-use for LLMs as an automated schema, which eliminates the need for hand-crafting task-specific demonstrations and carefully scripted interleaving of model generations with tool use.
\citet{DBLP:journals/corr/abs-2304-09842} harnesses the powerful decision-making abilities of LLMs, enabling them to combine various external tools to tackle compositional reasoning tasks. 

\begin{table*}[htb]
\centering
\footnotesize
\renewcommand\arraystretch{1.1}
\resizebox{\textwidth}{!}{
\begin{tabular}{l l cccc}
\toprule
\multirow{2}{*}{\textbf{Category}} & \multirow{2}{*}{\textbf{Representative Method}} & \multicolumn{4}{c}{\textbf{Comparison Scope}} \\ \cline{3-6} & & \textbf{Prompt Acquisition} & \textbf{Prompt Type} & \textbf{Language Model} & \textbf{Training Scenario} \\
\toprule
 & POTTER~\citep{DBLP:journals/corr/abs-2111-00539} & Manual & Template & BART/T5 & full fine-tune \\
 & CoT~\citep{DBLP:journals/corr/abs-2201-11903} & Manual & CoT & UL2/LaMDA/GPT-3 175B/Codex/PaLM & few-shot prompt \\
 & Auto-CoT~\citep{zhang2023automatic} & LM Generated & CoT & GPT-3 175B/Codex & few-shot prompt \\
 \multirow{-4}{*}{Prompt Engineering} & Least-to-Most~\citep{zhou2023leasttomost} & Manual & CoT & GPT-3 175B/Codex & few-shot prompt \\
\midrule
 & Calibrator~\citep{ye2022the} & Manual & Rationales & InstructGPT & few-shot fine-tune \\
 & Self-Consistency~\citep{DBLP:journals/corr/abs-2203-11171} & Manual & CoT & UL2/LaMDA/Codex/PaLM & few-shot prompt \\
 & DIVERSE~\citep{DBLP:journals/corr/abs-2206-02336} & LM Generated & CoT & GPT-3 175B/Codex & few-shot prompt \\
 \multirow{-4}{*}{Process Optimization} & LMSI~\citep{DBLP:journals/corr/abs-2210-11610} & LM Generated & CoT & PaLM & self-train \\
\midrule
 & PAL~\citep{DBLP:journals/corr/abs-2211-10435} & Manual & Code & Codex & few-shot prompt \\
 & PoT~\citep{DBLP:journals/corr/abs-2211-12588} & Manual & Code & Codex & few-shot prompt \\
\multirow{-3}{*}{External Engine} & Toolformer~\citep{DBLP:journals/corr/abs-2302-04761} & Manual & CoT with tools & GPT-J & self-train \\ 
\midrule
\midrule
 & RAINIER~\citep{liu-etal-2022-rainier} & LM Generated & Knowledge & UnifiedQA & few-shot prompt \\
 & PINTO~\citep{wang2023pinto} & LM Generated & Rationales & ROBERTA/T5 & full fine-tune \\
\multirow{-3}{*}{Implicit Knowledge} & Fine-tune-CoT~\citep{DBLP:journals/corr/abs-2212-10071} & LM Generated & Rationales & GPT-3 0.3B/1.3B/6.7B & full fine-tune \\
\midrule
 & PROMPTPG~\citep{lu2023dynamic} & Retrieval & CoT & GPT-3 175B & few-shot prompt \\
 \multirow{-2}{*}{Explicit Knowledge} & IRCoT~\citep{DBLP:journals/corr/abs-2212-10509} & Retrieval & CoT with wiki & Flan-T5/GPT-3 & few-shot prompt \\ 
\bottomrule
\end{tabular}
}
\caption{Comparison of reasoning with prompting methods from different scopes.}
\label{tab:prompt_comparison}
\end{table*}

\begin{figure} 
    \centering
    \resizebox{3in}{!}{
    \includegraphics{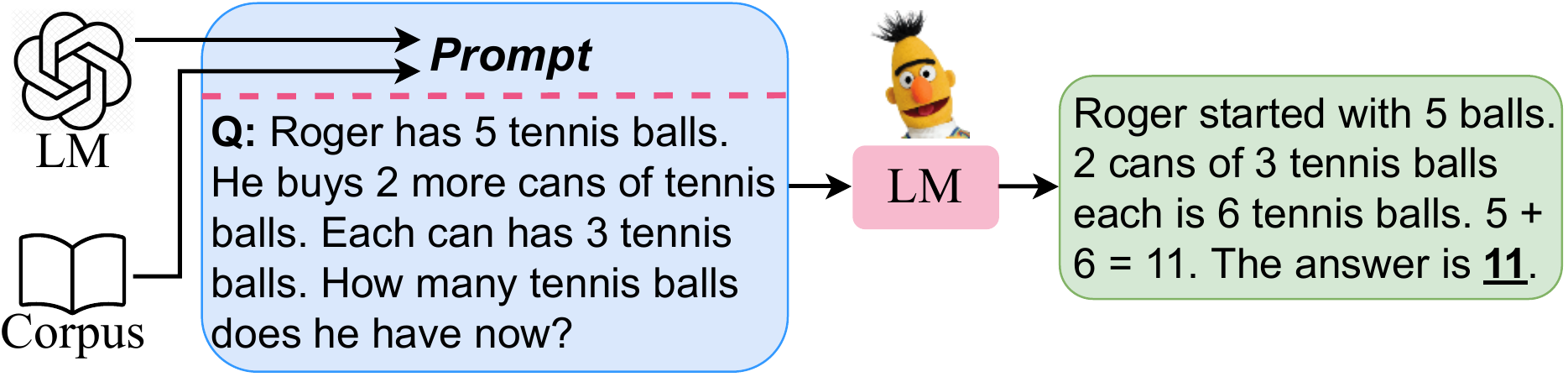}}
    \caption{Knowledge Enhanced Reasoning (\S \ref{knowledge}). 
    Prompts are generated by LM (\textbf{Implicit Knowledge}) or retrieved from external corpus (\textbf{Explicit Knowledge}).}
    \label{fig:knowledge_enhanced}
\end{figure}

\subsection{Knowledge Enhanced Reasoning}
\label{knowledge}

As noted in \citet{manning2022human}, knowledge plays a vital role in AI reasoning systems.
Knowledge enhanced methods aim to prompt LMs with \emph{implicit} (\S \ref{implicit_know}) or \emph{explicit} (\S \ref{explicit_know}) knowledge to assist in reasoning (see Figure~\ref{fig:knowledge_enhanced}).

\subsubsection{Implicit Knowledge}
\label{implicit_know}

Researchers have shown that LMs contain considerable implicit knowledge \citep{DBLP:conf/emnlp/DavisonFR19,DBLP:conf/emnlp/PetroniRRLBWM19,DBLP:journals/tacl/JiangXAN20}.
The following works try to induce such ``modeledge'' as knowledge-informed prompts for reasoning.

\citet{DBLP:conf/acl/0010LLWWBCH22} applies GPT-3 \citep{DBLP:conf/nips/BrownMRSKDNSSAA20} with few-shot prompting to generate knowledge and prompts the downstream LM. 
\citet{liu-etal-2022-rainier} draws support from reinforcement learning \citep{DBLP:journals/corr/SchulmanWDRK17} to further calibrate the knowledge. 
Different from the approaches using few-shot prompting in the knowledge generation stage, \citet{sun-etal-2022-tsgp} proposes a two-stage generative prompting which additionally includes answer generation prompts. 
Other works \citep{DBLP:journals/corr/abs-2210-06726,wang2023pinto,DBLP:journals/corr/abs-2212-00193,DBLP:journals/corr/abs-2212-08410,DBLP:journals/corr/abs-2212-10071} follow knowledge distillation that generates reasoning samples by prompting a larger LM and teaches smaller LMs.

\subsubsection{Explicit Knowledge}
\label{explicit_know}

Although large LMs have shown strong generation ability \citep{DBLP:conf/naacl/WiegreffeHSRC22,DBLP:journals/corr/abs-2210-06726,wang2023pinto}, they still have the tendency to hallucinate facts \citep{DBLP:conf/emnlp/RohrbachHBDS18} and generate inconsistent knowledge \citep{liu-etal-2022-rainier}. 
Recent works show that retrieving prompts for in-context learning is a nice means to achieve good performance \citep{DBLP:conf/acl-deelio/LiuSZDCC22,DBLP:conf/naacl/RubinHB22}.

Due to the instability of common retrieval approaches to measure the similarity of structured information, \citet{lu2023dynamic} proposes a dynamic prompt retrieval method based on policy gradient strategy, without brute-force searching. 
\citet{DBLP:journals/corr/abs-2301-00303} retrieves relevant knowledge based on the reasoning steps of CoT to provide more faithful explanations.
\citet{DBLP:journals/corr/abs-2212-10509} augments CoT prompting by persistently retrieving wiki documents for open-domain knowledge-intensive tasks that require complex multi-step reasoning.

\section{Comparison and Discussion}
\label{comparison}

\subsection{Comparison of Language Models}
\label{comparison_LM}
Table~\ref{tab:prompt_comparison} shows four comparison scopes of different methods. 
We further illustrate the performance comparison of LMs with different scales on GSM8K \citep{DBLP:journals/corr/abs-2110-14168} of arithmetic reasoning in Figure~\ref{fig:arithmetic}. 
Similar results on commonsense reasoning benchmarks are shown in Appendix~\ref{app:commmonsense}.

\begin{figure}
    \centering
    \resizebox{2.6in}{!}{
    \includegraphics{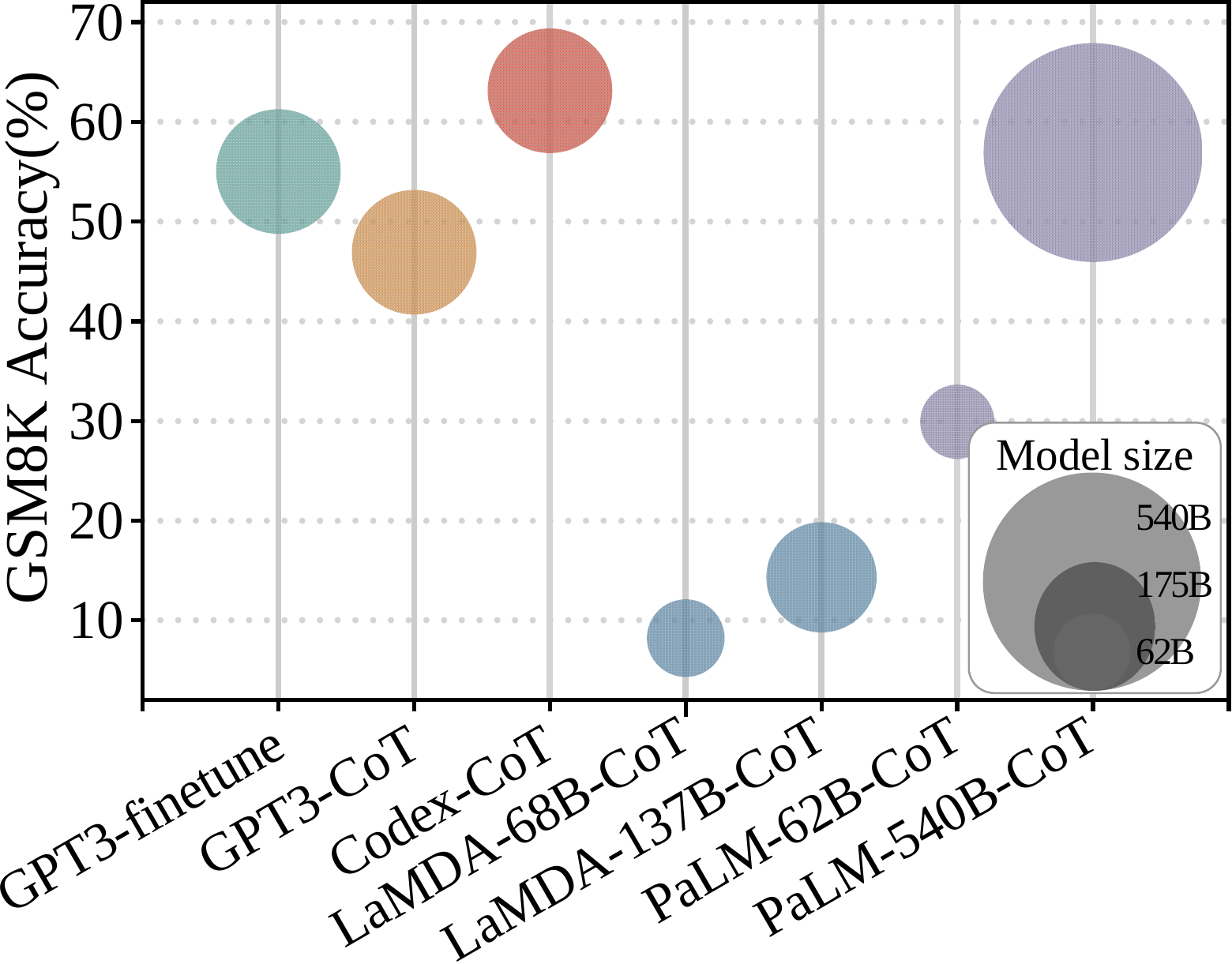}}
    \caption{
    Performance of different language model scales on arithmetic reasoning. Representatively, we show CoT \citep{DBLP:journals/corr/abs-2201-11903} experimental results on GSM8K \citep{DBLP:journals/corr/abs-2110-14168}.}
    \label{fig:arithmetic}
\end{figure}

\citet{DBLP:journals/corr/abs-2201-11903} systematically demonstrates that few-shot prompting performs better in almost all tasks as model scale increases, which can be explained by the fact that \textbf{LMs with larger model size contain more implicit knowledge for reasoning} \citep{DBLP:journals/corr/abs-2211-09110}.
Moreover, CoT prompting produces much greater increases, with PaLM-540B showing the greatest improvements, as depicted in Figure~\ref{fig:arithmetic}\&\ref{fig:commonsense}.
However, when the model scale declines to less than 100B, CoT prompting will yield no performance gain and may even be detrimental.
Thus, CoT prompting elicits an emergent ability of model scale \citep{wei2022emergent}.
One possibility is that when the stored knowledge reaches a certain level, the reasoning ability of LMs undergoes a qualitative change from quantitative change, leading to the emergence of emergent capabilities.
Additionally, \citet{DBLP:journals/corr/abs-2206-04615} points out that such ability generally occurs in multi-process tasks which may be explained that the evaluation only focuses on the final answer, but ignores the improvement of the middle process brought by the increase of model scale when it is not large enough.
Another intriguing observation is depicted in Figure~\ref{fig:arithmetic}\&\ref{fig:commonsense} that PaLM-62B \citep{DBLP:journals/corr/abs-2204-02311} even performs better than LaMDA-137B \citep{DBLP:journals/corr/abs-2201-08239}, possibly because it was trained on the higher-quality corpus.
This phenomenon leads us to speculate that such emergent ability is not solely determined by model parameter scale but also related to the quality of pre-training data.

Notably, Figure~\ref{fig:arithmetic}\&\ref{fig:commonsense} also illustrate that holding the same parameter scale, Codex \citep{DBLP:journals/corr/abs-2107-03374} outperforms GPT-3 significantly\footnote{Note that Codex and GPT-3 in our paper refer to code-davinci-002 and text-davinci-002 respectively in OpenAI API.}, even though the major difference between them is the training corpus (Codex is a GPT-3 variant training on code).
This phenomenon can also be inspected in recent works \citep{zhou2023leasttomost,DBLP:journals/corr/abs-2206-02336,zhang2023automatic,DBLP:journals/corr/abs-2210-07128,DBLP:journals/corr/abs-2211-09110}, indicating that \textbf{pre-training on code branch not only enables the ability of code generation/understanding but may also trigger the reasoning ability with CoT}. 
The exact cause is still elusive, but one intuition is that code is a form of text more similar to reasoning, thinking about procedure-oriented programming is analogous to solving problems step by step, and object-oriented programming is analogous to decomposing complex tasks into simpler ones \citep{fu2022gptroadmap}.
In addition, \citet{DBLP:journals/corr/abs-2304-03843} finds that CoT is beneficial only when the training data exhibits local structure. Due to its expertise in reasoning by navigating through multiple variables, CoT excels in deducing the relationship between two variables that have seldom been encountered in the same context. However, it may not perform better than simple statistical estimators when it comes to reasoning with variables that frequently co-occur in the training data.

\subsection{Comparison of Prompts}
\label{comparison_prompt}

Table~\ref{tab:prompt_comparison} shows the comparison of different methods of reasoning with LM prompting. 
There are three main sources of prompts for existing methods:
1) \textbf{Manual} construction is suitable for template-based prompts and few-shot prompting where the prompt is uncomplicated. 
2) \textbf{LM Generated} prompt makes up for the shortcomings of manual construction prompt. 
It can customize specific rationales for each question and provide sufficient knowledge with the prompt for fine-tuning or self-training. 
3) \textbf{Retrieval}-based prompt often relies on well-annotated external resources (e.g., Wikipedia) and consumes expensive information retrieval, but it can alleviate the unstable issue of the generation.

We observe that no matter how prompt is produced, CoT prompting only works on large LMs.
Smaller LMs work by fine-tuning with rationales.
Combined with the empirical conclusion in \citet{ye2022the}, these phenomena reveal that \textbf{high-quality reasoning rationales contained in the input context are the keys for reasoning with LM prompting}.
Although some works have attempted to explore the in-context learning ability on large LMs \citep{DBLP:conf/iclr/XieRL022,DBLP:journals/corr/abs-2202-12837,DBLP:journals/corr/abs-2211-15661}, the reason why CoT prompting can succeed is still intriguing to the community and not well-understood. 
One possible hypothesis is that CoT is a magical side product of training on code that can be unlocked by prompt.
Note that exemplars containing CoT in few-shot prompts can be viewed as a kind of instruction that arouses the reasoning ability hidden in large LMs.
\citet{DBLP:journals/corr/abs-2210-11416} verifies the similar result using CoT in instruction fine-tuning to advance model performance further.
In fact, in-context learning can be seen as an intermediate state of evolution from general prompts to human-readable instructions.
Following this trend, prompts may grow into an essential interface of human-machine interaction.

\section{Benchmarks and Resources}
\label{benchmark}

\subsection{Taxonomy of Benchmarks and Tasks}
\label{tasks}

In this section, we will give a brief overview of reasoning benchmarks and tasks. 
More details of datasets, as well as reasoning with ChatGPT can be found in Appendix~\ref{app:benchmark} and \ref{app:chatgpt}.

\paragraph{Arithmetic Reasoning.}
Arithmetic reasoning, also referred to as mathematical reasoning, is the ability to perform reasoning on \emph{math word problems} (MWP).
Early works on this task \citep{hosseini-etal-2014-learning,DBLP:conf/acl/KushmanZBA14,10.1162/tacl_a_00118,10.1162/tacl_a_00160,roy-roth-2015-solving} focus on relatively small datasets consisting of grade school single-step or multi-step MWP.
Later works increase in complexity, difficulty, and scale.
Most recently, \citet{DBLP:journals/corr/abs-2210-17517} extends existing datasets to construct a unified benchmark concerning mathematical abilities, language diversity, and external knowledge.

\paragraph{Commonsense Reasoning.}
Commonsense knowledge and commonsense reasoning are some of the major issues in machine intelligence \citep{DBLP:journals/corr/abs-1904-01172,DBLP:conf/aaai/Bhargava022}. 
When answering a question, people often draw upon their rich world knowledge.
For LMs, the major challenge of performing commonsense reasoning lies in how to involve physical and human interactions under the presumption of general background knowledge \citep{DBLP:conf/aaai/Bhargava022}.
Many benchmark datasets and tasks \citep{DBLP:journals/corr/abs-1803-05457,mihaylov-etal-2018-suit,DBLP:conf/naacl/TalmorHLB19,Bisk_Zellers_Lebras_Gao_Choi_2020,geva-etal-2021-aristotle} are designed, and the most widely used benchmark today is CommonsenseQA \citep{DBLP:conf/naacl/TalmorHLB19}.

\paragraph{Logical Reasoning.}
Common forms of logical reasoning include deductive reasoning and inductive reasoning, deductive reasoning and abductive reasoning \cite{sinha-etal-2019-clutrr,DBLP:conf/nesy/BaoPHTDWL22,DBLP:conf/acl/Young0BW22,DBLP:journals/corr/abs-2305-12599}.
Deductive reasoning is performed by going from general information to specific conclusions. Typical datasets in this field consist of synthetic rule bases plus derived conclusions \citep{ijcai2020p0537,tafjord-etal-2021-proofwriter}.
\citet{dalvi-etal-2021-explaining} creatively proposes a dataset containing multi-step entailment trees together with rules and conclusions.
As opposed to deductive reasoning, inductive reasoning aims to draw conclusions by going from specific observations to general principles \citep{DBLP:journals/corr/abs-2212-10923}. 

\paragraph{Symbolic Reasoning.}
Symbolic reasoning here only refers to a narrow collection of simple tasks that test a diverse set of symbolic manipulation functions, rather than symbolic AI, which is a more general concept.
Typical symbolic reasoning tasks include last letter concatenation, reverse list and coin flip \citep{DBLP:journals/corr/abs-2201-11903}.

\paragraph{Multimodal Reasoning.}  
Except for textual modality, humans utilize the information available across different modalities when performing reasoning.
To this end, multimodal reasoning benchmarks \citep{DBLP:conf/cvpr/ZellersBFC19,DBLP:conf/eccv/ParkBMFC20,dong-etal-2022-premise} are presented to narrow this gap. 
Recently, \citet{DBLP:journals/corr/abs-2209-09513} presents ScienceQA, a large-scale multimodal multiple choice dataset that consists of diverse questions of science topics with corresponding answers and explanations.
\citet{DBLP:journals/corr/abs-2210-00312} proposes a new task of multimodal analogical reasoning over knowledge graphs.

\subsection{Resources}
\label{resources}

Thanks to the open-source spirit of the NLP community, numerous resources are publicly available alongside papers for researchers to experiment with.
ThoughtSource is a central, open resource and community around data and tools related to CoT reasoning in large language models\footnote{\url{https://github.com/OpenBioLink/ThoughtSource}}.
The LangChain library is designed to help developers build applications using LLMs combined with other sources of computation or knowledge\footnote{\url{https://github.com/hwchase17/langchain}}.
$\lambda$prompt allows for building a complete large LM-based prompt machines, including ones that self-edit to correct and even self-write their own execution code\footnote{\url{https://github.com/approximatelabs/lambdaprompt}}. 
Recently, \citet{easyinstruct} develops EasyInstruct, a Python package for instructing LLMs like GPT-3 in research experiments.
A test case for reasoning using EasyInstruct can be found in Appendix~\ref{app:easyinstruct}.

\section{Future Directions}
\label{future_directions}


\paragraph{Theoretical Principle of Reasoning.} 

LMs have been demonstrated to have emergent zero-shot learning and reasoning abilities \citep{DBLP:journals/corr/abs-2201-11903,DBLP:journals/corr/abs-2203-11171,wei2022emergent}. 
To uncover the mystery of such a success, many researchers have empirically explored the role of in-context learning \citep{ye2022the,DBLP:conf/acl-deelio/LiuSZDCC22} and rationales \citep{DBLP:journals/corr/abs-2202-12837,DBLP:journals/corr/abs-2204-02329}.
Some works concentrate on unraveling the principles of machine learning algorithm  behind in-context learning \citep{DBLP:conf/iclr/XieRL022,DBLP:conf/iclr/AkyurekSA0Z23} and prompt tuning \citep{DBLP:conf/nips/WeiXM21}.
Another line of works tries to investigate the architecture of Transformers via knowledge neurons \cite{DBLP:conf/acl/DaiDHSCW22} or skill neurons \cite{DBLP:journals/corr/abs-2211-07349}. 
More recent works \citep{DBLP:journals/corr/abs-2210-12810,DBLP:journals/corr/abs-2210-07128} demonstrate that pre-trained LMs of code are better handling structured commonsense reasoning and prediction than LMs of natural language, even when the downstream task does not involve source code at all.
However, the code-based pre-training (or re-structured pre-training \cite{DBLP:journals/corr/abs-2206-11147}) still has limitations since it has to utilize off-the-shelf structure (e.g., existing aligned corpus or build from scratch via syntax tree or AMR \cite{DBLP:conf/acllaw/BanarescuBCGGHK13}) to reformulate plain texts.
Thus, the truth may be close, and we argue that it is beneficial to study the theoretical principle to advocate for a transparent view of reasoning with LM prompting and further decipher the dark matter of intelligence by highlighting the counterintuitive continuum across language, knowledge, and reasoning\footnote{Keynote talk on ACL 2022 entitled ``2082: An ACL Odyssey: The Dark Matter of Intelligence and Language''.}.
Note that reasoning in NLP has the potential advantages of complex problem-solving and should better utilize dark matters in cross-disciplines (e.g., Theory of Mind~\cite{DBLP:journals/corr/abs-2210-13312,DBLP:journals/corr/abs-2304-11490,DBLP:journals/corr/abs-2212-10060,shapira2023clever}). 

\paragraph{Efficient Reasoning.} 
To be noted, existing methods mainly depend on large LMs, which may consume high computing resources. 
Regarding practicality, it is necessary to study reasoning with small LMs or develop efficient reasoning methodologies which pay attention to carbon emission and energy usage during model training and inference \cite{DBLP:journals/corr/abs-2111-05193}.
One feasible way may be developing models that can enable generalization across a range of evaluation scenarios such as  Flan-T5 \citep{DBLP:journals/corr/abs-2210-11416}, which finetune both with and without exemplars (i.e., zero-shot and few-shot) and with and without CoT.
Recently, an intuitive approach has been proposed to transfer the reasoning capabilities of large LMs to smaller LMs via knowledge distillation \citep{DBLP:journals/corr/abs-2212-00193,DBLP:journals/corr/abs-2212-08410,DBLP:journals/corr/abs-2212-10071}.
Other promising directions include retrieval augmentation \citep{DBLP:journals/corr/abs-2202-01110}, model editing \citep{DBLP:conf/emnlp/CaoAT21,DBLP:conf/iclr/MitchellLBFM22,DBLP:conf/icml/MitchellLBMF22,DBLP:journals/corr/abs-2301-10405}, delta-tuning \citep{DBLP:conf/iclr/HeZMBN22,DBLP:conf/acl/MaoMHAM0YK22,DBLP:conf/acl-dialdoc/PalKR22,DBLP:journals/corr/abs-2203-06904}, etc.

\paragraph{Robust, Faithful and Interpretable Reasoning.} 
Robustness, faithfulness and interpretability have long been pursued by the field of deep learning, especially in tasks that require strong logic, like reasoning. 
\citet{DBLP:journals/corr/abs-2212-08061} demonstrates that zero-shot CoT will produce undesirable toxicity and biases, indicating the necessity of robust, faithful and interpretable reasoning. 
\citet{DBLP:journals/corr/abs-2208-14271} leverages a selection-inference \cite{DBLP:journals/corr/abs-2205-09712} multi-stage architecture for faithful reasoning, but there is still a lack of interpretability within each stage. 
Code-based works \cite{DBLP:journals/corr/abs-2210-07128,DBLP:journals/corr/abs-2211-10435,DBLP:journals/corr/abs-2211-12588} reach robustness and interpretability to some extent, but they have the aid of an external engine.
There is still a long way to achieve true robustness, faithfulness and interpretability with LMs. 
Fortunately, \citet{DBLP:journals/corr/abs-2207-10342} provides a new idea for utilizing a probabilistic program to tackle various reasoning problems. 
Other solutions may be neural-symbolic approaches \cite{DBLP:conf/acl/DuDX0020,DBLP:conf/acl/Li0CSQ22,DBLP:journals/corr/abs-2105-10334,DBLP:journals/tacl/FengYZG22} or human feedback \citep{DBLP:journals/corr/abs-2203-02155}.

\paragraph{Multimodal (Interactive) Reasoning.} 
Textual reasoning is restricted to what can be expressed through natural language.
A more promising direction is multimodal reasoning regarding the information diversity of the real world of human reasoning. 
\citet{DBLP:journals/corr/abs-2209-09513} generates CoT when dealing with a multimodal dataset; however, it simply extracts textual descriptions from images, and it is still a textual reasoning task indeed.
Intuitively, it is beneficial to integrate multimodal information into reasoning processes such as images, audio, videos, etc., and design a unified multimodal CoT.
Apart from unified multimodal models, it is also promising to model chains \citep{DBLP:conf/chi/WuJD0MTC22} to conduct interactive reasoning among models of different modalities.
Besides, \citet{DBLP:journals/corr/abs-2210-13312} shows that one of today’s largest language models (GPT-3 \cite{DBLP:conf/nips/BrownMRSKDNSSAA20}) lacks the skill to reason about the mental states, and reactions of all people involved. 
Thus, interactive reasoning methodologies should be noted by inspiring from other domains (e.g., Cognitive Science \cite{DBLP:journals/corr/abs-1904-02682}, Social Intelligence \cite{doi:10.1073/pnas.2115730119}), which may have potential guidance for reasoning in NLP since only increasing the scale of LMs is likely not the most effective way to create AI systems. 

\paragraph{Generalizable (True) Reasoning.} 

Generalization is one of the most significant symbols of models to attain true reasoning abilities. 
Given a reasoning task, we hope LMs can handle not only the problem itself but solve a group of similar reasoning tasks (not seen during training). 
\citet{DBLP:journals/corr/abs-2211-09066,DBLP:journals/corr/abs-2207-04901} explore the OOD problem on the length of reasoning questions, but the true generalization is still far from satisfactory. 
Meanwhile, \citet{DBLP:journals/natmi/KejriwalSMM22} highlights that more comprehensive evaluation methods grounded in theory (e.g., naive physics \cite{DBLP:journals/ai/GardinM89} and commonsense psychology \cite{DBLP:journals/aim/GordonH04}) should be proposed.
We argue that the generalizable reasoning may be closely related to analogy reasoning \citep{DBLP:conf/acl/ChenXFSLZSLXZ22,DBLP:journals/corr/abs-2212-09196}, causal reasoning \citep{DBLP:journals/tacl/FederKMPSWEGRRS22}, compositional reasoning \citep{DBLP:conf/naacl/YangJYZYY22}, etc.

\section{Conclusion and Vision}

In this paper, we provide a review of reasoning with language model prompting, including comprehensive comparisons, and several research directions.
In the future, we envision a more potent synergy between the methodologies from the NLP and other domains and hope sophisticated and efficient LM prompting models will increasingly contribute to improving reasoning performance.

\section*{Acknowledgment}

We would like to express gratitude to the anonymous reviewers for their kind comments.
We thank Yicong for correcting an inappropriate Equation in the paper.
This work was supported by the National Natural Science Foundation of China (No.62206246 and U19B2027), Zhejiang Provincial Natural Science Foundation of China (No. LGG22F030011), Ningbo Natural Science Foundation (2021J190), and Yongjiang Talent Introduction Programme (2021A-156-G), CAAI-Huawei MindSpore Open Fund, and NUS-NCS Joint Laboratory (A-0008542-00-00).
The computing resources were supported by the Information Technology Center and State Key Lab of CAD\&CG, Zhejiang University.

 
\section*{Limitations}

In this study, we provide a survey of reasoning with language model prompting.
We discuss the related surveys in Appendix \ref{related-survey} and will continue adding more related approaches with more detailed analysis.
Despite our best efforts, there may be still some limitations that remain in this paper.

\paragraph{References \& Methods.} Due to the page limit, we may miss some important references 
and cannot afford all the technical details.
We mainly review the cutting-edge methods within two years (mostly in 2022) in \S \ref{categories}, mainly from the ACL, EMNLP, NAACL, NeurIPS, ICLR, arXiv, etc., and we will continue to pay attention to and supplement the latest works.

\paragraph{Benchmarks.}
Most of the reasoning benchmarks mentioned in \S \ref{benchmark} are gathered and categorized from the experimental part of mainstream works.
The definition and boundary of each task may not be accurate enough.
Besides, our work may miss some kind of reasoning tasks such as reasoning with generics \cite{DBLP:journals/corr/abs-2205-11658}, default inheritance reasoning \cite{DBLP:conf/ijcai/Brewka87}, non-monotonic reasoning \cite{ginsberg1987readings} in NLP, and will try our best to fulfill this gap.

\paragraph{Empirical Conclusions.}
We give detailed comparisons and discussions of language models and prompts in \S \ref{comparison}, and list some promising future directions in \S \ref{future_directions}.
All the conclusions are proposed and further speculated upon empirical analysis of existing works which may not be macroscopic enough.
As the field evolves faster, we will update the latest opinions timely.




\bibliography{custom}
\bibliographystyle{acl_natbib}

\appendix
\section{Appendix}
\label{sec:appendix}

\tikzstyle{my-box}=[
    rectangle,
    draw=hidden-draw,
    rounded corners,
    text opacity=1,
    minimum height=1.5em,
    minimum width=5em,
    inner sep=2pt,
    align=center,
    fill opacity=.5,
]
\tikzstyle{leaf}=[my-box, minimum height=1.5em,
    fill=hidden-orange!60, text=black, align=left,font=\scriptsize,
    inner xsep=2pt,
    inner ysep=4pt,
]
\begin{figure*}[tp]
    \centering
    \resizebox{\textwidth}{!}{
        \begin{forest}
            forked edges,
            for tree={
                grow=east,
                reversed=true,
                anchor=base west,
                parent anchor=east,
                child anchor=west,
                base=left,
                font=\small,
                rectangle,
                draw=hidden-draw,
                rounded corners,
                align=left,
                minimum width=4em,
                edge+={darkgray, line width=1pt},
                s sep=3pt,
                inner xsep=2pt,
                inner ysep=3pt,
                ver/.style={rotate=90, child anchor=north, parent anchor=south, anchor=center},
            },
            where level=1{text width=3em,font=\scriptsize,}{},
            where level=2{text width=5.6em,font=\scriptsize,}{},
            where level=3{text width=5.5em,font=\scriptsize,}{},
            where level=4{text width=6.1em,font=\scriptsize,}{},
            [
                Reasoning with Language Model Prompting, ver
                [
                    Taxonomy \\ of Methods \\ (\S \ref{categories})
                    [
                        Strategy Enhanced \\ Reasoning (\S \ref{stategy})
                        [
                            Prompt Engineering \\ (\S \ref{prompt_engineering})
                            [
                                Single-Stage
                                [
                                    Contrastive~\cite{DBLP:conf/acl/ParanjapeMGHZ21}{,}
                                    POTTER~\cite{DBLP:journals/corr/abs-2111-00539}{,}
                                    CoT~\cite{DBLP:journals/corr/abs-2201-11903}{,} \\ 
                                    ZeroCoT~\cite{DBLP:journals/corr/abs-2205-11916}{,}
                                    Complexity~\cite{fu2023complexitybased}{,}
                                    Multilingual~\cite{DBLP:journals/corr/abs-2210-03057}{,} \\ 
                                    Auto-CoT~\cite{zhang2023automatic}{,}
                                    Table~\cite{DBLP:journals/corr/abs-2210-06710}{,}
                                    AlgoPrompt ~\cite{DBLP:journals/corr/abs-2211-09066}{,} \\ 
                                    Active-Prompt~\cite{DBLP:journals/corr/abs-2302-12246}{,}
                                    Automate-CoT~\cite{DBLP:journals/corr/abs-2302-12822}
                                    , leaf, text width=25em
                                ]
                            ]
                            [
                                Multi-Stage
                                [
                                    iCAP~\cite{DBLP:journals/corr/abs-2203-08383}{,}
                                    SI~\cite{DBLP:journals/corr/abs-2205-09712}{,}
                                    Least-to-Most~\cite{zhou2023leasttomost}{,} \\
                                    MAIEUTIC~\cite{jung-etal-2022-maieutic}{,}
                                    Faithful~\cite{DBLP:journals/corr/abs-2208-14271}{,}
                                    Decomposed \\ ~\cite{khot2023decomposed}{,}
                                    Self-Ask~\cite{DBLP:journals/corr/abs-2210-03350}{,}
                                    Successive~\cite{dua-etal-2022-successive}{,}
                                    LMLP \\ ~\cite{zhang2022the}{,}
                                    LAMBADA~\cite{DBLP:journals/corr/abs-2212-13894}{,}
                                    Iter-Decomp~\cite{DBLP:journals/corr/abs-2301-01751}
                                    , leaf, text width=25em
                                ]
                            ]
                        ]
                        [
                            Process Optimization \\ (\S \ref{path_optimization})
                            [
                                Self-Optimization
                                [
                                    Calibrator~\cite{ye2022the}{,}
                                    Human-AI~\cite{DBLP:conf/naacl/WiegreffeHSRC22}
                                    , leaf, text width=25em
                                ]
                            ]
                            [
                                Ensemble-Optimization
                                [
                                    Self-C~\cite{DBLP:journals/corr/abs-2203-11171}{,}
                                    DIVERSE~\cite{DBLP:journals/corr/abs-2206-02336}{,}
                                    Complexity~\cite{fu2023complexitybased}{,} \\
                                    Self-V~\cite{DBLP:journals/corr/abs-2212-09561}{,}
                                    MCR~\cite{yoran2023answering}
                                    , leaf, text width=25em
                                ]
                            ]
                            [
                                Iterative-Optimization
                                [
                                    STaR~\cite{DBLP:journals/corr/abs-2203-14465}{,}
                                    LMSI~\cite{DBLP:journals/corr/abs-2210-11610}{,} \\ 
                                    Reflexion~\cite{DBLP:journals/corr/abs-2303-11366}{,}
                                    Self-Refine~\cite{DBLP:journals/corr/abs-2303-17651}{,}
                                    REFINER~\cite{DBLP:journals/corr/abs-2304-01904}
                                    , leaf, text width=25em
                                ]
                            ]
                        ]
                        [
                            External Engine \\ (\S \ref{external_engine})
                            [
                                Physical Simulator
                                [
                                    Mind's Eye~\cite{liu2023minds}
                                    , leaf, text width=25em
                                ]
                            ]
                            [
                                Code Interpreter
                                [
                                    COCOGEN~\cite{DBLP:journals/corr/abs-2210-07128}{,}
                                    PAL~\cite{DBLP:journals/corr/abs-2211-10435}{,}
                                    PoT~\cite{DBLP:journals/corr/abs-2211-12588}{,}
                                    \\ 
                                    Faithful-CoT~\cite{DBLP:journals/corr/abs-2301-13379}{,}
                                    Versa-Decomp~\cite{DBLP:journals/corr/abs-2301-13808}{,}
                                    SynPrompt \\ ~\cite{DBLP:journals/corr/abs-2302-00618}{,}
                                    MathPrompter~\cite{imani2023mathprompter}
                                    , leaf, text width=25em
                                ]
                            ]
                            [
                                Tool Learning
                                [
                                    Toolformer~\cite{DBLP:journals/corr/abs-2302-04761}{,}
                                    ART~\cite{DBLP:journals/corr/abs-2303-09014}{,}
                                    Chameleon~\cite{DBLP:journals/corr/abs-2304-09842}
                                    , leaf, text width=25em
                                ]
                            ]
                        ]
                    ]
                    [
                        Knowledge Enhanced \\ Reasoning (\S \ref{knowledge})
                        [
                            Implicit Knowledge \\ (\S \ref{implicit_know})
                            [
                                GenKnow~\cite{DBLP:conf/acl/0010LLWWBCH22}{,}
                                RAINIER~\cite{liu-etal-2022-rainier}{,}
                                MT-CoT~\cite{DBLP:journals/corr/abs-2210-06726}{,}
                                PINTO~\cite{wang2023pinto}{,}
                                TSGP \\ ~\cite{sun-etal-2022-tsgp}{,} 
                                DecompDistill~\cite{DBLP:journals/corr/abs-2212-00193}{,}
                                Teaching~\cite{DBLP:journals/corr/abs-2212-08410}{,}
                                Fine-tune-CoT~\cite{DBLP:journals/corr/abs-2212-10071}{,} \\ 
                                Specializing~\cite{DBLP:journals/corr/abs-2301-12726}
                                , leaf, text width=32.7em
                            ]
                        ]
                        [
                            Explicit Knowledge \\ (\S \ref{explicit_know})
                            [
                                LogicSolver~\cite{yang-etal-2022-logicsolver}{,}
                                Vote-\emph{k}~\cite{su2023selective}{,}
                                PROMPTPG~\cite{lu2023dynamic}{,}
                                IRCoT~\cite{DBLP:journals/corr/abs-2212-10509}{,} \\ 
                                RR~\cite{DBLP:journals/corr/abs-2301-00303}
                                , leaf, text width=32.7em
                            ]
                        ]
                    ]
                ]
                [
                    Taxonomy \\ of Tasks \\(\S \ref{benchmark})
                    [
                        Arithmetic
                        [
                            CoT~\cite{DBLP:journals/corr/abs-2201-11903}{,}
                            Self-C~\cite{DBLP:journals/corr/abs-2203-11171}{,}
                            STaR~\cite{DBLP:journals/corr/abs-2203-14465}{,}
                            LogicSolver~\cite{yang-etal-2022-logicsolver}{,}
                            Least-to-Most \\ ~\cite{zhang2022the}{,}
                            ZeroCoT~\cite{DBLP:journals/corr/abs-2205-11916}{,}
                            DIVERSE~\cite{DBLP:journals/corr/abs-2206-02336}{,}
                            Minerva~\cite{DBLP:journals/corr/abs-2206-14858}{,}
                            PROMPTPG \\ ~\cite{lu2023dynamic}{,}
                            Complexity~\cite{fu2023complexitybased}{,}
                            Multilingual~\cite{DBLP:journals/corr/abs-2210-03057}{,}
                            Auto-CoT~\cite{zhang2023automatic}{,}
                            LMSI~\cite{DBLP:journals/corr/abs-2210-11610}{,} \\ 
                            AlgoPrompt~\cite{DBLP:journals/corr/abs-2211-09066}{,}
                            PAL~\cite{DBLP:journals/corr/abs-2211-10435}{,}
                            PoT~\cite{DBLP:journals/corr/abs-2211-12588}{,}
                            DecompDistill~\cite{DBLP:journals/corr/abs-2212-00193}{,}
                            LMP \\ ~\cite{DBLP:journals/corr/abs-2212-06094}{,}
                            Teaching~\cite{DBLP:journals/corr/abs-2212-08410}{,}
                            Self-V~\cite{DBLP:journals/corr/abs-2212-09561}{,}
                            Fine-tune-CoT~\cite{DBLP:journals/corr/abs-2212-10071}{,}
                            Specializing \\ ~\cite{DBLP:journals/corr/abs-2301-12726}{,}
                            Faithful-CoT~\cite{DBLP:journals/corr/abs-2301-13379}{,}
                            Toolformer~\cite{DBLP:journals/corr/abs-2302-04761}{,}
                            SynPrompt~\cite{DBLP:journals/corr/abs-2302-00618}{,}
                            Active-Prompt~\cite{DBLP:journals/corr/abs-2302-12246}{,} \\ 
                            Automate-CoT~\cite{DBLP:journals/corr/abs-2302-12822}{,}
                            MathPrompter~\cite{imani2023mathprompter}{,}
                            ART~\cite{DBLP:journals/corr/abs-2303-09014}{,}
                            REFINER~\cite{DBLP:journals/corr/abs-2304-01904}{,} \\ 
                            Self-Refine~\cite{DBLP:journals/corr/abs-2303-17651}
                            , leaf, text width=39.7em
                        ]
                    ]
                    [
                        Commonsense
                        [
                            Contrastive~\cite{DBLP:conf/acl/ParanjapeMGHZ21}{,}
                            POTTER~\cite{DBLP:journals/corr/abs-2111-00539}{,}
                            CoT~\cite{DBLP:journals/corr/abs-2201-11903}{,}
                            GenKnow~\cite{DBLP:conf/acl/0010LLWWBCH22}{,} 
                            Self-C \\ ~\cite{DBLP:journals/corr/abs-2203-11171}{,} 
                            STaR~\cite{DBLP:journals/corr/abs-2203-14465}{,}
                            Calibrator~\cite{ye2022the}{,}
                            MAIEUTIC~\cite{jung-etal-2022-maieutic}{,}
                            ZeroCoT~\cite{DBLP:journals/corr/abs-2205-11916}{,} \\
                            DIVERSE~\cite{DBLP:journals/corr/abs-2206-02336}{,}
                            Vote-\emph{k}~\cite{su2023selective}{,}
                            RAINIER~\cite{liu-etal-2022-rainier}{,}
                            Self-Ask~\cite{DBLP:journals/corr/abs-2210-03350}{,} 
                            Auto-CoT~\cite{zhang2023automatic}{,} \\
                            Human-AI~\cite{DBLP:conf/naacl/WiegreffeHSRC22}{,}
                            MT-CoT~\cite{DBLP:journals/corr/abs-2210-06726}{,}
                            COCOGEN~\cite{DBLP:journals/corr/abs-2210-07128}{,} 
                            LMSI~\cite{DBLP:journals/corr/abs-2210-11610}{,}
                            PINTO \\ ~\cite{wang2023pinto}{,} 
                            TSGP~\cite{sun-etal-2022-tsgp}{,}
                            Teaching~\cite{DBLP:journals/corr/abs-2212-08410}{,}
                            Fine-tune-CoT~\cite{DBLP:journals/corr/abs-2212-10071}{,}
                            IRCoT~\cite{DBLP:journals/corr/abs-2212-10509}{,} \\ 
                            RR~\cite{DBLP:journals/corr/abs-2301-00303}{,}
                            Faithful-CoT~\cite{DBLP:journals/corr/abs-2301-13379}{,}
                            Active-Prompt~\cite{DBLP:journals/corr/abs-2302-12246}{,}
                            Automate-CoT~\cite{DBLP:journals/corr/abs-2302-12822}{,} \\ 
                            Reflexion~\cite{DBLP:journals/corr/abs-2303-11366}{,}
                            MCR~\cite{yoran2023answering}
                            , leaf, text width=39.7em
                        ]
                    ]
                    [
                        Logical
                        [
                            SI~\cite{DBLP:journals/corr/abs-2205-09712}{,}
                            Faithful~\cite{DBLP:journals/corr/abs-2208-14271}{,}
                            LMLP~\cite{zhang2022the}{,} \\ 
                            Self-V~\cite{DBLP:journals/corr/abs-2212-09561}{,}
                            LAMBADA~\cite{DBLP:journals/corr/abs-2212-13894}{,}
                            Faithful-CoT~\cite{DBLP:journals/corr/abs-2301-13379}{,}
                            , leaf, text width=39.7em
                        ]
                    ]
                    [
                        Symbolic
                        [
                            CoT~\cite{DBLP:journals/corr/abs-2201-11903}{,}
                            Self-C~\cite{DBLP:journals/corr/abs-2203-11171}{,}
                            STaR~\cite{DBLP:journals/corr/abs-2203-14465}{,}
                            Least-to-Most~\cite{zhang2022the}{,} \\
                            ZeroCoT~\cite{DBLP:journals/corr/abs-2205-11916}{,}
                            Decomposed~\cite{khot2023decomposed}{,}
                            Auto-CoT~\cite{zhang2023automatic}{,}
                            PAL~\cite{DBLP:journals/corr/abs-2211-10435}{,} \\ 
                            Teaching~\cite{DBLP:journals/corr/abs-2212-08410}{,}
                            Fine-tune-CoT~\cite{DBLP:journals/corr/abs-2212-10071}{,}
                            SynPrompt~\cite{DBLP:journals/corr/abs-2302-00618}{,}
                            Active-Prompt~\cite{DBLP:journals/corr/abs-2302-12246}{,} \\ 
                            Automate-CoT~\cite{DBLP:journals/corr/abs-2302-12822}{,}
                            ART~\cite{DBLP:journals/corr/abs-2303-09014}
                            , leaf, text width=39.7em
                        ]
                    ]
                    [
                        Multimodal
                        [
                            ScienceQA~\cite{DBLP:journals/corr/abs-2209-09513}{,}
                            MarT~\cite{DBLP:journals/corr/abs-2210-00312}{,}
                            IPVR~\cite{https://doi.org/10.48550/arxiv.2301.05226}{,}
                            Multimodal-CoT~\cite{DBLP:journals/corr/abs-2302-00923}{,} \\ 
                            KOSMOS-1~\cite{DBLP:journals/corr/abs-2302-14045}{,}
                            Visual-ChatGPT~\cite{DBLP:journals/corr/abs-2303-04671}{,}
                            ViperGPT~\cite{DBLP:journals/corr/abs-2303-08128}{,}
                            MM-REACT~\cite{DBLP:journals/corr/abs-2303-11381}{,} \\ 
                            Chameleon~\cite{DBLP:journals/corr/abs-2304-09842}
                            , leaf, text width=39.7em
                        ]
                    ]
                ]
            ]
        \end{forest}
    }
    \caption{Taxonomy of Reasoning with Language Model Prompting.}
    \label{fig:categorization_of_reasoning_big}
\end{figure*}

\subsection{Related Survey}
\label{related-survey}
As this area is relatively nascent, only a few surveys exist.
Closest to our work, \citet{DBLP:journals/corr/abs-2212-10403} gives a survey towards reasoning with large language models.
\citet{DBLP:journals/corr/abs-2301-00234} organizes and discusses the advanced techniques of in-context learning.
\citet{DBLP:journals/corr/abs-2303-18223} reviews the latest advancements in Large Language Models (LLMs) and delves into the unresolved challenges that will shape future developments.
\citet{DBLP:conf/aaai/Bhargava022} covers methods for commonsense knowledge reasoning and generation with pre-trained LMs.
\citet{DBLP:journals/corr/abs-2212-10535} reviews the key tasks, datasets, and methods at the intersection of mathematical reasoning and deep learning over the past decade.
\citet{DBLP:journals/corr/abs-2212-05767} surveys knowledge graph reasoning tracing from static to temporal and then to multi-modal knowledge graphs.
\citet{DBLP:journals/corr/abs-2302-07842} reviews works in which language models (LMs) are augmented with reasoning skills and the ability to use tools.
\citet{DBLP:journals/corr/abs-2202-12205} conducts a survey of studies implementing neural-symbolic (NeSy) NLP approaches for reasoning and so on.
\citet{DBLP:journals/corr/abs-2206-05675} provides a survey of several popular works dealing with uncertainty reasoning.
\citet{DBLP:journals/corr/abs-2304-08354} concentrates on the leverage of external tools by LLMs which is also called Tool Learning.
Other surveys focusing on prompt learning \cite{DBLP:journals/corr/abs-2107-13586} or pre-trained models \cite{DBLP:journals/corr/abs-2003-08271,DBLP:conf/ijcai/DuLLZ22} are also related to our work.

Unlike those surveys, in this paper, we conduct a review of reasoning with LM prompting, hoping to systematically understand the methodologies, compare different methods and inspire new ideas.

\subsection{Taxonomy of Methods and Tasks}
\label{app:taxonomy}

We list the complete taxonomy of reasoning with language model prompting from methods and tasks in Figure~\ref{fig:categorization_of_reasoning_big}.

\subsection{Performance Comparison of LMs with Different Scales}
\label{app:commmonsense}

\begin{figure}[ht]
    \centering
    \resizebox{3in}{!}{
    \includegraphics{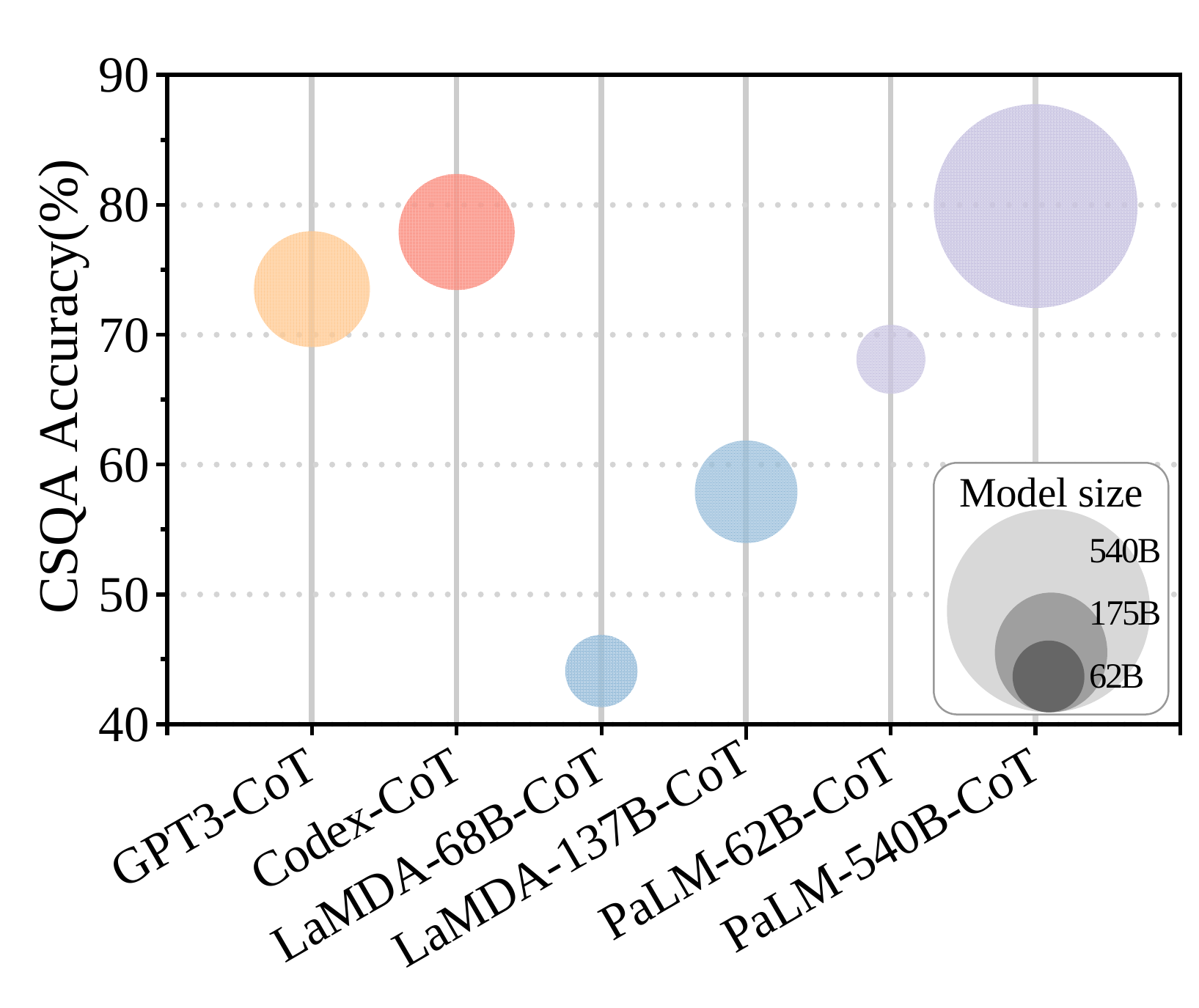}}
    \caption{Performance of different language model scales on commonsense reasoning. Representatively, We show CoT \citep{DBLP:journals/corr/abs-2201-11903} experimental results on CommonsenseQA \citep{DBLP:conf/naacl/TalmorHLB19}.}
    \label{fig:commonsense}
\end{figure}

To show the generalization of discussions in \S \ref{comparison_LM} on different reasoning tasks, we additionally show the performance comparison of LMs with different scales on CommonsenseQA \citep{DBLP:conf/naacl/TalmorHLB19} of commonsense reasoning in Figure~\ref{fig:commonsense}.

\subsection{Detailed Information of Reasoning Benchmarks}
\label{app:benchmark}

In \S~\ref{benchmark}, we give a brief overview
on benchmarks and tasks requiring various reasoning
skills.
We list more benchmarks and show their key statistics in Table~\ref{tab:benchmarks}.
Apart from the above-mentioned specific reasoning tasks in \S~\ref{benchmark}, there are some benchmarks \citep{DBLP:journals/corr/abs-1711-00350,DBLP:journals/corr/abs-2206-04615,DBLP:journals/corr/abs-2212-08286} that can evaluate the model's more diverse and generalized reasoning capabilities, which can also be included in the category of reasoning tasks.

\begin{table*}[htb]
\centering
\footnotesize
\begin{tabular}{llcccc}
\toprule
 \multicolumn{1}{l}{\multirow{2}[2]{*}{\textbf{Task}}} & \multirow{2}[2]{*}{\textbf{Dataset}} & \multicolumn{4}{c}{\textbf{Size}}  \\
        \cmidrule(lr){3-6}
          &       & \multicolumn{1}{c}{\textbf{Train}} & \multicolumn{1}{c}{\textbf{Valid}} & \multicolumn{1}{c}{\textbf{Test}} & \multicolumn{1}{c}{\textbf{All}} \\
\midrule
 & AddSub \citep{hosseini-etal-2014-learning} & 395 & - & -  & 395 \\       
 & SingleOp \citep{10.1162/tacl_a_00118} & 562 & - & - & 562  \\
 & SingleEq \citep{10.1162/tacl_a_00160} & 508 & - & - & 508 \\
 & MultiArith \citep{roy-roth-2015-solving} & 600 & -  & - & 600 \\
 & Dophin18k \citep{huang-etal-2016-well} & 18,460 & - & - & 18,460 \\
 & MAWPS \citep{koncel-kedziorski-etal-2016-mawps} & 1,921 & - & - & 1,921 \\
 & Math23k \citep{wang-etal-2017-deep} & 23,161 & - & - & 23,161 \\
 & AQUA-RAT \citep{ling-etal-2017-program} & 97,467 & - & 254 & 97,721  \\
 & MathQA \citep{amini-etal-2019-mathqa}  & 29,807 & 4,471 & 2,981 & 37,259 \\
 & DROP \citep{dua-etal-2019-drop} & 5,850 & - & - & 5,850 \\ 
 & ASDiv \citep{miao-etal-2020-diverse} & 1,217 & - & - & 1,217 \\
& GSM8K \citep{DBLP:journals/corr/abs-2110-14168} & 7,473 & - &  1,319 & 8,792  \\
 & SVAMP \citep{patel-etal-2021-nlp} & 1,000 & - & - & 1,000 \\
 & MATH \citep{DBLP:conf/nips/HendrycksBKABTS21} & 7,500 & - & 5,000 & 12,500 \\
 & NumGLUE \citep{mishra-etal-2022-numglue} & 101,835 & - & - & 101,835 \\
\multirow{-15}{*}{Arithmetic Reasoning} & Lila \citep{DBLP:journals/corr/abs-2210-17517} & 133,815 & - & - & 133,815 \\
\midrule
 & Last Letter Concatenation \citep{DBLP:journals/corr/abs-2201-11903} & - & - & - & - \\
  & Coin Flip \citep{DBLP:journals/corr/abs-2201-11903} & - & - & - & - \\
\multirow{-3}{*}{Symbolic Reasoning} & Reverse List \citep{DBLP:journals/corr/abs-2201-11903}  & - & - & - & - \\
\midrule
 & ARC \citep{DBLP:journals/corr/abs-1803-05457} & 3,370 & 869 & 3,548 & 7,787 \\
 & OpenBookQA \citep{mihaylov-etal-2018-suit} & 4,957 & 500 & 500 & 5,957 \\
 & CommonsenseQA \citep{DBLP:conf/naacl/TalmorHLB19} & 9,741 & 1,221 & 1,140 & 12,102 \\
 & PIQA \citep{Bisk_Zellers_Lebras_Gao_Choi_2020} & 16,000 & 2,000 & 3,000  & 21,000 \\
\multirow{-5}{*}{Commonsense Reasoning} & StrategyQA \citep{geva-etal-2021-aristotle} & 2,290 & - & 490 & 2,780 \\
\midrule
 & RuleTaker \citep{ijcai2020p0537} & 14,135 & 2,019 & 3,038 & 20,192 \\
 & ProofWriter \citep{tafjord-etal-2021-proofwriter} & - & - & - & - \\
 & EntailmentBank \citep{dalvi-etal-2021-explaining} & 1,313 & 187 & 340 & 1,840 \\
 & CLUTRR \citep{sinha-etal-2019-clutrr} &  6,016 & - & - &  6,016 \\
\multirow{-5}{*}{Logical Reasoning} & DEER  \citep{DBLP:journals/corr/abs-2212-10923} &  1,200 & - & - &  1,200 \\
\midrule
 & VCR \citep{DBLP:conf/cvpr/ZellersBFC19} & 212,923   & 26,534& 25,263 & 264,720 \\
 & VisualCOMET \citep{DBLP:conf/eccv/ParkBMFC20} & 1,174,063 & 146,332 & 145,309 & 1,465,704 \\
 & VLEP \citep{lei-etal-2020-likely} & 20,142 & 4,392 & 4,192 & 28,726 \\
 & PMR \citep{dong-etal-2022-premise} & 12,080 & 1,538 & 1,742 & 15,360 \\
\multirow{-5}{*}{Multimodal Reasoning} & ScienseQA \citep{DBLP:journals/corr/abs-2209-09513} & 12,726 & 4,241 & 4,241 & 21,208 \\
\bottomrule
\end{tabular}
\caption{An overview of benchmarks and tasks on reasoning. }
\label{tab:benchmarks}
\end{table*}

\subsection{Reasoning with ChatGPT}
\label{app:chatgpt}

\begin{figure}[ht]
    \centering
    \resizebox{.48\textwidth}{!}{
    \includegraphics{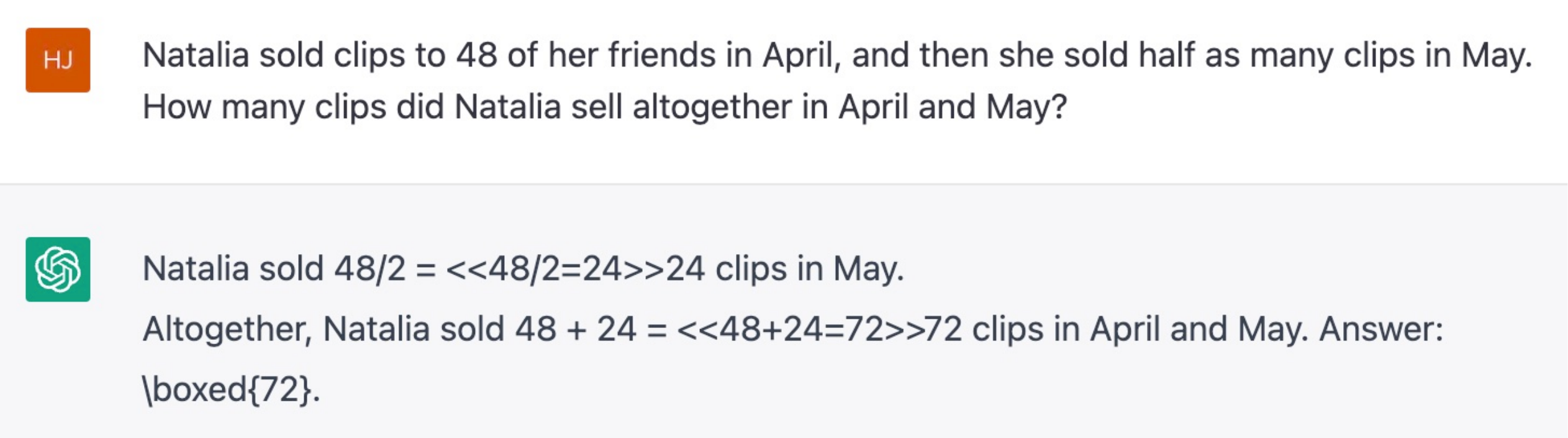}}
    \caption{A test case from GSM8K \citep{DBLP:journals/corr/abs-2110-14168} on ChatGPT \citep{ChatGPT-OpenAI}.}
    \label{fig:chatgpt_math}
\end{figure}

\begin{figure}[ht]
    \centering
    \resizebox{.48\textwidth}{!}{
    \includegraphics{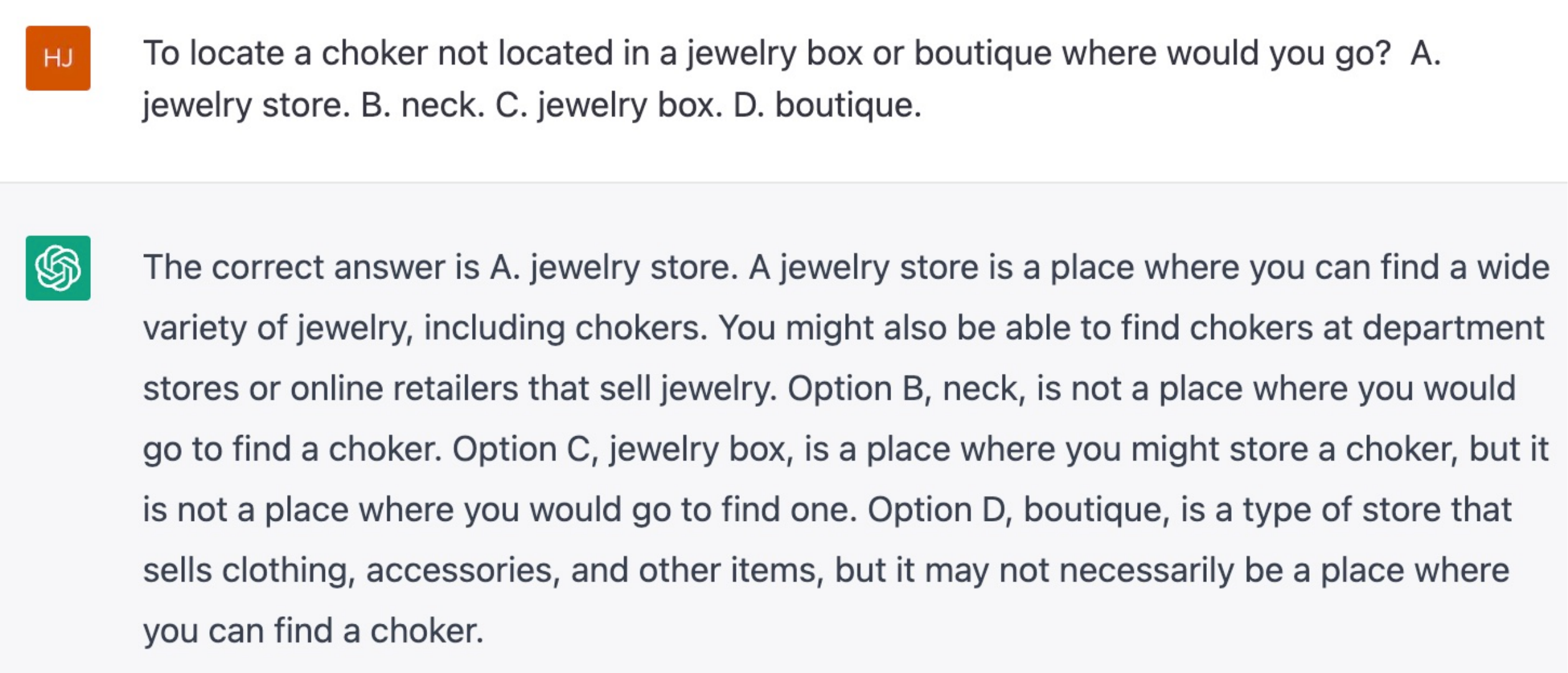}}
    \caption{A test case from CommonsenseQA \citep{DBLP:conf/naacl/TalmorHLB19} on ChatGPT \citep{ChatGPT-OpenAI}.}
    \label{fig:chatgpt_commonsense}
\end{figure}

\begin{figure}[ht]
    \centering
    \resizebox{.48\textwidth}{!}{
    \includegraphics{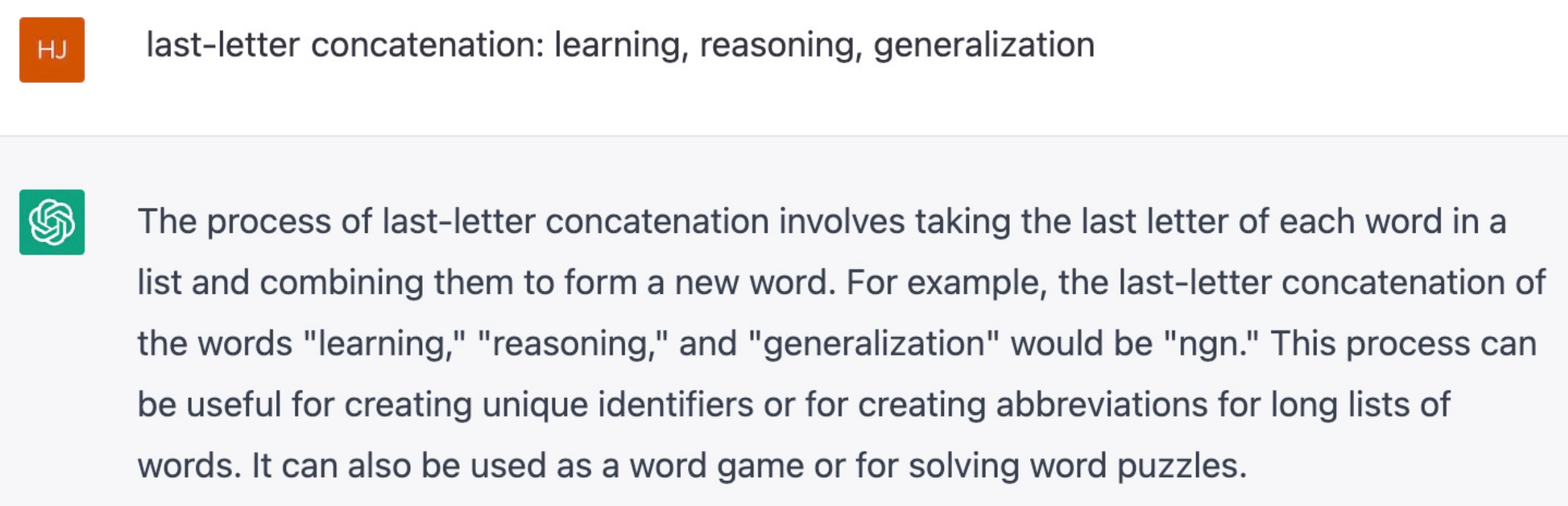}}
    \caption{A test case from Last Letter Concatenation \citep{DBLP:journals/corr/abs-2201-11903} on ChatGPT \citep{ChatGPT-OpenAI}.}
    \label{fig:chatgpt_symbolic}
\end{figure}

Recently, \citet{ChatGPT-OpenAI} develops ChatGPT, an AI chatbot system that has attracted tremendous users. ChatGPT is trained on a massive dataset of text and is able to generate human-like responses to a wide variety of prompts, the promising approach for which is called Reinforcement Learning from Human Feedback \citep{DBLP:journals/corr/abs-2203-02155}.
The backbone of ChatGPT is from a model in the GPT-3.5 large LM series\footnote{\url{https://beta.openai.com/docs/model-index-for-researchers}}.
In order to savor the reasoning ability of large LMs more realistically, we conduct some case tests on ChatGPT.
Concretely, we pick out a piece of data from GSM8K \citep{DBLP:journals/corr/abs-2110-14168}, CommonsenseQA \citep{DBLP:conf/naacl/TalmorHLB19} and Last Letter Concatenation \citep{DBLP:journals/corr/abs-2201-11903} which respectively represent arithmetic reasoning, commonsense reasoning, and symbolic reasoning.
Then we test each of the selected data on ChatGPT directly.
Results can be seen in Figure~\ref{fig:chatgpt_math}-\ref{fig:chatgpt_symbolic}.

Figure~\ref{fig:chatgpt_math} shows that given a math problem in GSM8K \citep{DBLP:journals/corr/abs-2110-14168}, ChatGPT outputs a reasoning process and a correct answer without in-context exemplars.
This blazes its powerful arithmetic reasoning ability.
The reasoning process has the same format as the gold label in GSM8K, indicating that GSM8K may be contained in the training corpus of ChatGPT.

In Figure~\ref{fig:chatgpt_commonsense}, we test ChatGPT on a piece of data in CommsonsenseQA \citep{DBLP:conf/naacl/TalmorHLB19}.
It not only gives the correct answer but additionally details why each option is right or wrong, which does not appear in the gold label of the dataset.
This demonstrates the strong commonsense reasoning ability of ChatGPT.

Figure~\ref{fig:chatgpt_symbolic} is a case in Last Letter Concatenation \citep{DBLP:journals/corr/abs-2201-11903}. We observe that although ChatGPT gives a detailed and accurate description of last letter concatenation, it fails to answer the given question, showing that its symbolic reasoning capability is not as excellent as the above two.

\subsection{Reasoning using EasyInstruct}
\label{app:easyinstruct}

\begin{figure}[ht]
    \centering
    \resizebox{.48\textwidth}{!}{
    \includegraphics{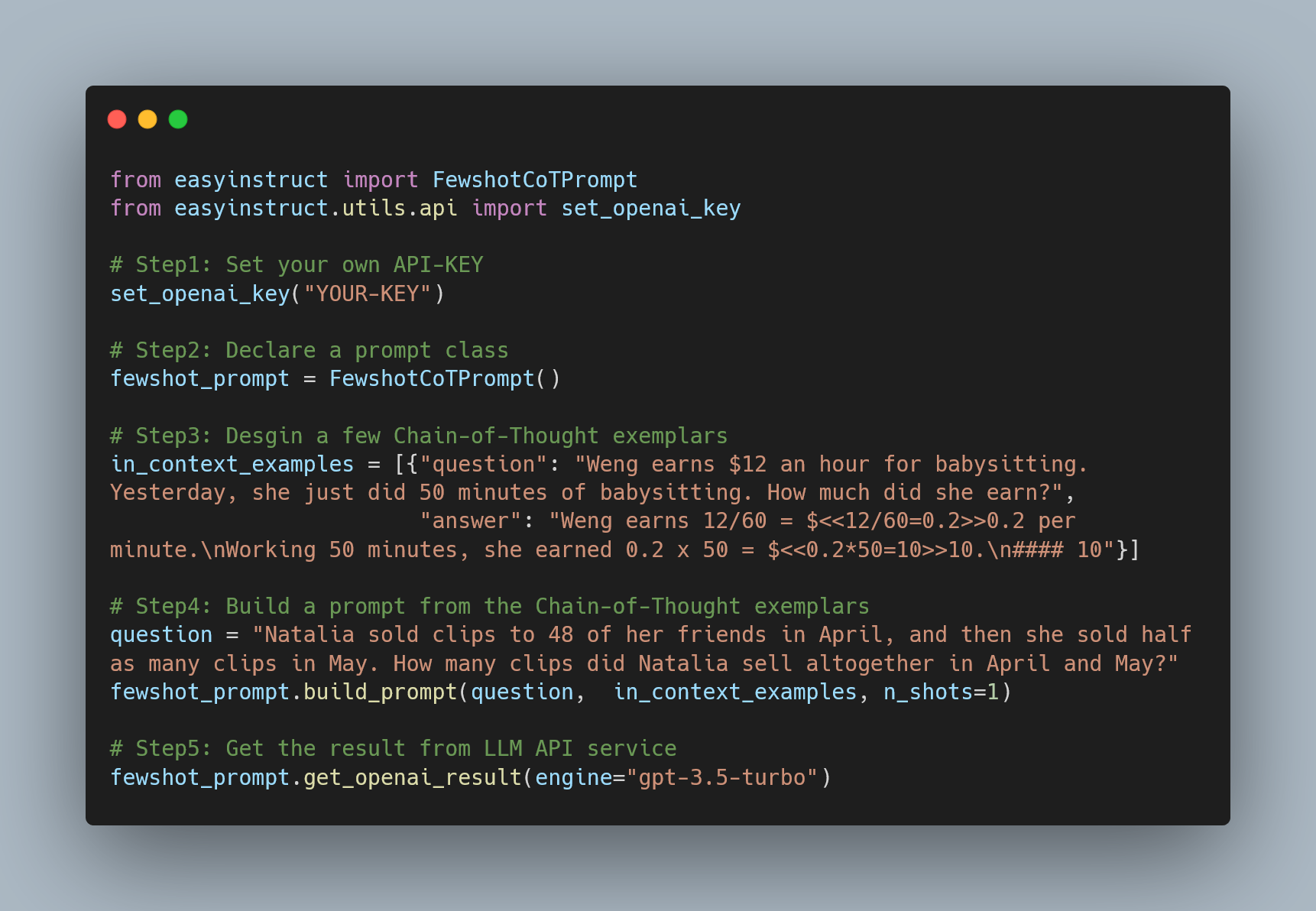}}
    \caption{A test case from GSM8K \citep{DBLP:journals/corr/abs-2110-14168} using EasyInstruct \citep{easyinstruct}.}
    \label{fig:easyinstruct}
\end{figure}

\end{document}